\documentclass[sigconf]{acmart}

\usepackage{booktabs}
\usepackage{bm}
\usepackage{url}
\usepackage{mathrsfs}
\usepackage{multirow}
\usepackage{balance}
\usepackage{amsthm}
\usepackage{amsmath}
\usepackage{amstext}
\usepackage[ruled,vlined]{algorithm2e}
\usepackage{subcaption}
\usepackage{overpic}
\usepackage{color}
\usepackage{enumitem}
\usepackage{fancyvrb}
\usepackage{graphicx}
\usepackage{caption}
\usepackage{multirow}
\usepackage{bm}

\newtheorem{definition}{Definition}
\newtheorem{theorem}{Theorem}

\newcommand{\indep}{\perp \!\!\! \perp}
\newcommand{\slfrac}[2]{\left.#1\middle/#2\right.}

\DeclareMathOperator*{\argmin}{arg\,min}

\AtBeginDocument{%
  \providecommand\BibTeX{{%
    \normalfont B\kern-0.5em{\scshape i\kern-0.25em b}\kern-0.8em\TeX}}}

\copyrightyear{2021}
\acmYear{2021}
\setcopyright{acmcopyright}\acmConference[KDD '21]{Proceedings of the 27th ACM SIGKDD Conference on Knowledge Discovery and Data Mining}{August 14--18, 2021}{Virtual Event, Singapore}
\acmBooktitle{Proceedings of the 27th ACM SIGKDD Conference on Knowledge Discovery and Data Mining (KDD '21), August 14--18, 2021, Virtual Event, Singapore}
\acmPrice{15.00}
\acmDOI{10.1145/3447548.3467333}
\acmISBN{978-1-4503-8332-5/21/08}

\begin{document}

\fancyhead{}

\title{Model-Based Counterfactual Synthesizer for Interpretation}

\author{Fan Yang$^1$, Sahan Suresh Alva$^1$, Jiahao Chen$^2$, Xia Hu$^1$}
\affiliation{
  \institution{$^1$Department of Computer Science and Engineering, Texas A\&M University, College Station, TX, USA} 
  \institution{$^2$J.P. Morgan AI Research, New York, NY, USA}
  }
 \email{{nacoyang, sahanalva, xiahu}@tamu.edu} 
 \email{jiahao.chen@jpmchase.com} 

\renewcommand{\shortauthors}{F. Yang et al.}

\begin{abstract}
  \textit{Counterfactuals}, serving as one of the emerging type of model interpretations, have recently received attention from both researchers and practitioners. Counterfactual explanations formalize the exploration of ``what-if'' scenarios, and are an instance of example-based reasoning using a set of hypothetical data samples. Counterfactuals essentially show how the model decision alters with input perturbations. Existing methods for generating counterfactuals are mainly algorithm-based, which are time-inefficient and assume the same counterfactual universe for different queries. To address these limitations, we propose a \textbf{\underline{M}}odel-based \textbf{\underline{C}}ounterfactual \textbf{\underline{S}}ynthesizer (MCS) framework for interpreting machine learning models. We first analyze the model-based counterfactual process and construct a base synthesizer using a conditional generative adversarial net (CGAN). To better approximate the counterfactual universe for those rare queries, we novelly employ the umbrella sampling technique to conduct the MCS framework training. Besides, we also enhance the MCS framework by incorporating the causal dependence among attributes with model inductive bias, and validate its design correctness from the causality identification perspective. Experimental results on several datasets demonstrate the effectiveness as well as efficiency of our proposed MCS framework, and verify the advantages compared with other alternatives. 
\end{abstract}

\ccsdesc[500]{Computing methodologies~Causal reasoning and diagnostics}
\ccsdesc[500]{Computing methodologies~Neural networks}
\ccsdesc[500]{Computing methodologies~Generative and developmental approaches}
\ccsdesc[300]{Computing methodologies~Supervised learning by classification}

\keywords{Counterfactual sample; causal explanation; model interpretation}

\maketitle

\section{Introduction}

Recently, machine learning (ML) models have been widely deployed in many real-world applications, and achieved a huge success in many domains. For high-stake scenarios, such as medical diagnosis~\cite{litjens2017survey} and policy making~\cite{brennan2013emergence}, explaining the behaviors of ML models is necessary for humans to be able to scrutinize the model outcomes before making a final decision. Various interpretation techniques~\cite{du2019techniques} have thus been proposed to handle the ML explainability issue. Feature attribution~\cite{du2019techniques}, for example, is one commonly used interpretation technique in many ML tasks, including image captioning, machine translation and question answering, where important features that contribute most are highlighted as evidences for model predictions.

However, most ML interpretations like feature attribution typically lack the reasoning capability, since they are not discriminative in nature~\cite{dhurandhar2018explanations}, which makes them limited in understanding \textit{how}. To further investigate the decision boundaries of ML models, \textit{counterfactual} explanation~\cite{wachter2017counterfactual}, emerging as a new form of interpretation, has gradually raised attentions in recent years. Counterfactuals are essentially a series of hypothetical samples, which are synthesized within a certain data distribution and can flip the model decisions to preferred outcomes. With valid counterfactuals, humans can understand how model predictions change with particular input perturbations, and further conduct reasoning under "what-if" circumstances. Given a loan rejection case for instance, attribution analysis simply indicates those important features for rejection (e.g., applicant income), while counterfactuals are able to show how the application could be approved with certain changes (e.g., increase the annual income from $\$50,000$ to $\$80,000$).

Several initial attempts on generating counterfactuals have been made in recent work. The first, and most straightforward, line of methodology is to utilize the adversarial perturbations to synthesize the hypothetical samples~\cite{moore2019explaining,mothilal2020explaining}, where the search process is conducted in the input data space. One significant drawback of this methodology is that out-of-distribution (OOD) samples cannot be avoided, which can largely limit the reasoning capability of generated counterfactuals. To cope with this issue, another line of methodology proposes to synthesize the counterfactuals in the latent code space with the aid of pre-trained generative models~\cite{joshi2018xgems,pawelczyk2020learning,yang2021generative}. In this way, the generated counterfactuals are then well guaranteed to be within certain data distribution for interpretation purposes. Besides, considering the huge computational complexity for deriving high-dimensional samples (e.g., images), there is also a line of methodology using feature composition in data space to synthesize counterfactuals~\cite{goyal2019counterfactual,vermeire2020explainable}. The extra requirement for this methodology is a proper distractor instance (could be the query itself), which provides relevant feature resources for synthesis.

Despite existing efforts, effective generation of counterfactuals still remains challenging. First, common counterfactual frameworks are mostly algorithm-based ones, which makes them inefficient for sample generation, because each new query necessitates solving one specific optimization problem at one time. For cases with multiple inputs, algorithm-based frameworks could be extremely time-consuming. Second, the counterfactual universe is unknown in advance, and its high-quality approximation is preferred for better explanation. Existing frameworks mostly assume the same counterfactual universe for different queries, though this may not be consistent with the settings of real-world counterfactuals which are related to the inputs. Third, causal dependence among attributes needs be considered for counterfactual feasibility. Although there are a few work~\cite{karimi2020survey,mahajan2019preserving} trying to incorporate causal constraints into counterfactuals, they simply achieve it by adding extra regularization terms, and do not have any theoretical guarantees on the generation process from a causality perspective.

To address the aforementioned challenges, we propose a \textbf{\underline{M}}odel-based \textbf{\underline{C}}ounterfactual \textbf{\underline{S}}ynthesizer (MCS) framework, which can faithfully capture the counterfactual universe and properly incorporate the attribute causal dependence. Specifically, by analyzing the counterfactual process, we are motivated to employ conditional generative adversarial net (CGAN)~\cite{mirza2014conditional} as the base, and further build a model-based synthesizer by introducing relevant counterfactual objectives. To make MCS better approximate the potential counterfactual universe, we novelly apply the umbrella sampling~\cite{kastner2011umbrella} in synthesizer training, aiming to properly consider the influence of those rare events in data on counterfactual reasoning. 
Moreover, we also use model inductive bias to design the generator architecture in our proposed MCS framework, so as to incorporate the causal dependence of attributes into the generated samples, which is further validated from the causality identification perspective. Our main contributions are summarized as follows:
\begin{itemize}[leftmargin=*]
\item Design a model-based counterfactual explanation framework (i.e., MCS) based on CGAN, whose goal is to help humans better understand the decision boundaries of deployed ML models; 

\item Apply the umbrella sampling technique in MCS training, which significantly enhances the synthesizer in capturing the influence of those rare events in data for counterfactual explanation;  

\item Use the concept of model inductive bias to design the generator architecture in the proposed MCS, and further validate the design correctness through a causality identification process;  

\item Demonstrate the advantages of our proposed MCS on different datasets, and compare the performance with other alternatives. 
\end{itemize}

\section{Preliminaries} 

In this section, we briefly introduce some involved concepts, as well as some basics of the employed techniques. 

\textbf{Counterfactual Explanation.}
This is one particular ML interpretation technique developed from example-based reasoning~\cite{rissland1991example}, where hypothetical data samples are provided to promote the understandings of model boundaries. As a specific example, consider a classification model $f_{\boldsymbol\theta}: \mathbb{R}^{d} \rightarrow \{-1,1\}$, with $-1$ and $1$ respectively denoting the undesired and desired outputs. The counterfactual explanation problem can be generally formulated as: 
\begin{equation}\label{eq_cf}
    \mathbf{x}^{*}=\argmin_{\mathbf{x}\sim\mathcal{C}} \ l(\mathbf{x}, \mathbf{q}_{0}) \quad
    \mathrm{s.t.} \ f_{\boldsymbol\theta}(\mathbf{q}_{0})=-1, \  f_{\boldsymbol\theta}(\mathbf{x}^{*})=1,
\end{equation}
where $\mathbf{q}_{0}$ represents the input query, and $\mathbf{x}^{*}$ is the derived counterfactual sample. Here, $\mathcal{C}$ indicates the counterfactual universe of the observed data space $\mathbb{R}^{d}$, and $l: \mathbb{R}^{d} \times \mathbb{R}^{d} \rightarrow \mathbb{R}^{+}_{0}$ denotes a distance measure in the input space. From Eq.~\ref{eq_cf}, we can see that counterfactuals are essentially data samples within some distributions, which can flip the model decisions as desired, while keeping similar to the query input. Conceptually, Eq.~\ref{eq_cf} can be solved either in an algorithm-based way, or a model-based way. Algorithm-based methods typically employ different optimization strategies to solve Eq.~\ref{eq_cf} for each query $\mathbf{q}_{0}$, while model-based ones try to approximate the particular $\mathcal{C}$ given relevant constraints and further conduct sampling. In this paper, we mainly explore the counterfactual explanation problem using the model-based methods.

\textbf{Generative Modeling with CGAN.} 
Generative adversarial net is a novel way to train generative models, which typically has a generator $G$ and a discriminator $D$~\cite{goodfellow2014generative}. The training objective of $G$ is to capture the data distribution, while the objective of $D$ is to estimate the probability that a sample comes from the data rather than $G$. CGAN is a natural extension of this framework, where $G$ and $D$ are both conditioned on some additional information $\mathbf{a}$ (e.g., labels or attributes). The min-max game between $G$ and $D$ conducted in CGAN training can be expressed as: 
\begin{equation}\label{eq_cgan}
    \min_{G} \max_{D} V(D, G) = \!\!\!\! \mathop{\mathbb{E}}_{\mathbf{x}\sim\mathcal{P}_{\mathbf{x}}} \!\!\! \log D(\mathbf{x}|\mathbf{a}) \ + \!\!
    \mathop{\mathbb{E}}_{\mathbf{z}\sim\mathcal{P}_{\mathbf{z}}} \!\!\! \log (1\!-\!D(G(\mathbf{z}|\mathbf{a}))),
\end{equation}
where $\mathcal{P}_{\mathbf{x}}$ indicates the data distribution over $\mathbf{x}$, $\mathcal{P}_{\mathbf{z}}$ denotes a prior noise distribution, and $V$ represents a value function of the two players in the min-max game. With a well-trained CGAN, we can effectively capture the conditional distribution given certain constraints or regularizations. In this paper, we use the CGAN framework to approximate the potential counterfactual universe.

\textbf{Model Inductive Bias.}
When multiple decisions are equally good, model inductive bias enables the learning process to prioritize some decisions over the others~\cite{mitchell1980need}, which is independent of the data observed. Model inductive bias can be incorporated in different ways. In early days, connectionist models commonly indicated their inductive bias through relevant regularization terms~\cite{mcclelland1992interaction}. For conventional Bayesian models, inductive bias is typically expressed through the prior distribution~\cite{griffiths2010probabilistic}, either from its selection or parameterization. In other contexts, inductive bias can also be encoded with model architectures~\cite{battaglia2018relational,yang2020deep}, where the structure itself indicates the data-generating assumption or the potential decision space. In this paper, we make use of the inductive bias specifically instilled by generator architecture to properly consider the causal dependence among attributes for generated samples.

\section{Model-Based Counterfactual Synthesizer Framework} 

In this section, we first analyze the counterfactual universe given a deployed ML model, and then formulate the problem of model-based counterfactual explanation. Further, we introduce the proposed synthesizer design based on the CGAN framework.

\subsection{Model-Based Counterfactual Explanation}

\begin{figure}
\centering
\includegraphics[width=0.70\columnwidth]{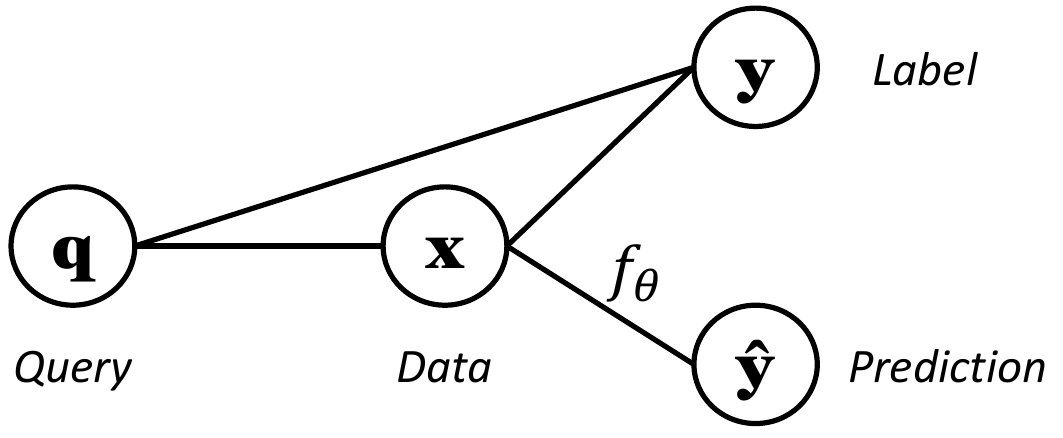}
\vspace{-0.1cm}
\caption{The general graphical model for counterfactual generation process given a deployed ML model.} 
\label{fig:cfgraph}
\end{figure} 

To effectively design a counterfactual synthesizer, it is crucial to make clear the counterfactual universe we are focusing on. Given a deployed ML model $f_{\boldsymbol\theta}$, we can characterize the whole universe with the graphical model illustrated in Fig.~\ref{fig:cfgraph}. In general, $\mathbf{x}$ and $\mathbf{y}$ represent the data and label variables respectively, while $\hat{\mathbf{y}}$ denotes the decision variable output from $f_{\boldsymbol\theta}$. The query variable $\mathbf{q}$ is introduced to incorporate relevant constraints for counterfactual reasoning, which brings about the fact that hypothetical samples (i.e., $\mathbf{x}$ and $\mathbf{y}$) are typically generated under the influence from $\mathbf{q}$. According to the graphical model in Fig.~\ref{fig:cfgraph}, we can further factorize the joint distribution of the whole counterfactual universe as follows:
\begin{equation}\label{eq_factor}
    \mathcal{P}_{\mathbf{q},\mathbf{x},\mathbf{y},\hat{\mathbf{y}}} = 
    \mathcal{P}_{\mathbf{q}} \cdot \mathcal{P}_{\mathbf{x}|\mathbf{q}} \cdot \mathcal{P}_{\hat{\mathbf{y}}|\mathbf{x}} \cdot \mathcal{P}_{\mathbf{y}|\mathbf{x},\mathbf{q}} 
    = \mathcal{P}_{\mathbf{q}} \cdot \mathcal{P}_{\hat{\mathbf{y}}|\mathbf{x}} \cdot \mathcal{P}_{\mathbf{x},\mathbf{y}|\mathbf{q}} \ .
\end{equation}
Within Eq.~\ref{eq_factor}, $\mathcal{P}_{\mathbf{q}}$ is typically known as the prior, and $\mathcal{P}_{\hat{\mathbf{y}}|\mathbf{x}}$ is considered as fixed since the model $f_{\boldsymbol\theta}$ is pre-deployed. Thus, the key to capturing the counterfactual universe lies in the proper approximation of $\mathcal{P}_{\mathbf{x},\mathbf{y}|\mathbf{q}}$ (i.e., the joint distribution of $\mathbf{x}$ and $\mathbf{y}$ conditioned on $\mathbf{q}$), which reflects the latent sample generation process with certain query. Thus, to achieve the model-based counterfactual analysis, we need to investigate the hypothetical sample generation under particular query conditions. 
With this insight, we now formally define the problem of model-based counterfactual explanation below.

\begin{definition}
  A \textbf{model-based counterfactual} is a data point sampled from a perturbed \textbf{hypothetical distribution}, which statistically satisfies the counterfactual requirements (indicated by Eq.~\ref{eq_cf}). Given a specific query $\mathbf{q}_{0}$, counterfactual $\mathbf{x}^{*}$ can be obtained through sampling $\mathbf{x}^{*}\sim\mathcal{C}_{\mathbf{x}|\mathbf{q}_{0}}$, where $\mathcal{C}_{\mathbf{x}|\mathbf{q}_{0}}$ is a hypothetical distribution marginalized from $\mathcal{C}_{\mathbf{x},\mathbf{y}|\mathbf{q}_{0}}$. In general, $\mathcal{C}_{\mathbf{x},\mathbf{y}|\mathbf{q}_{0}}$ can be derived by
  \begin{equation}
      \mathcal{C}_{\mathbf{x},\mathbf{y}|\mathbf{q}_{0}} = \argmin_{\mathcal{P}_{\mathbf{x},\mathbf{y}|\mathbf{q}_{0}}} \ L^{cf}(\mathcal{P}_{\mathbf{x},\mathbf{y}|\mathbf{q}_{0}}), 
  \end{equation}
  where $\mathbf{q}_{0}$ follows the prior $\mathcal{P}_{\mathbf{q}}$, and $L^{cf}$ indicates a counterfactual loss. 
\end{definition}

By definition, a model-based counterfactual does not focus on the instance optimization for each individual query. Instead, it tries to capture the latent sample generation process with particular query conditions, within the whole counterfactual universe. One significant merit brought by model-based explanations is that it largely enhances the efficiency for counterfactual generation, since we only need to obtain the certain hypothetical distribution once for all potential queries following the prior $\mathcal{P}_{\mathbf{q}}$. Nevertheless, modeling such latent generation processes is nontrivial, and we need some specific designs to make it effectively work.

\subsection{Conditional Synthesizer Design} 

\begin{figure}
\centering
\includegraphics[width=0.90\columnwidth]{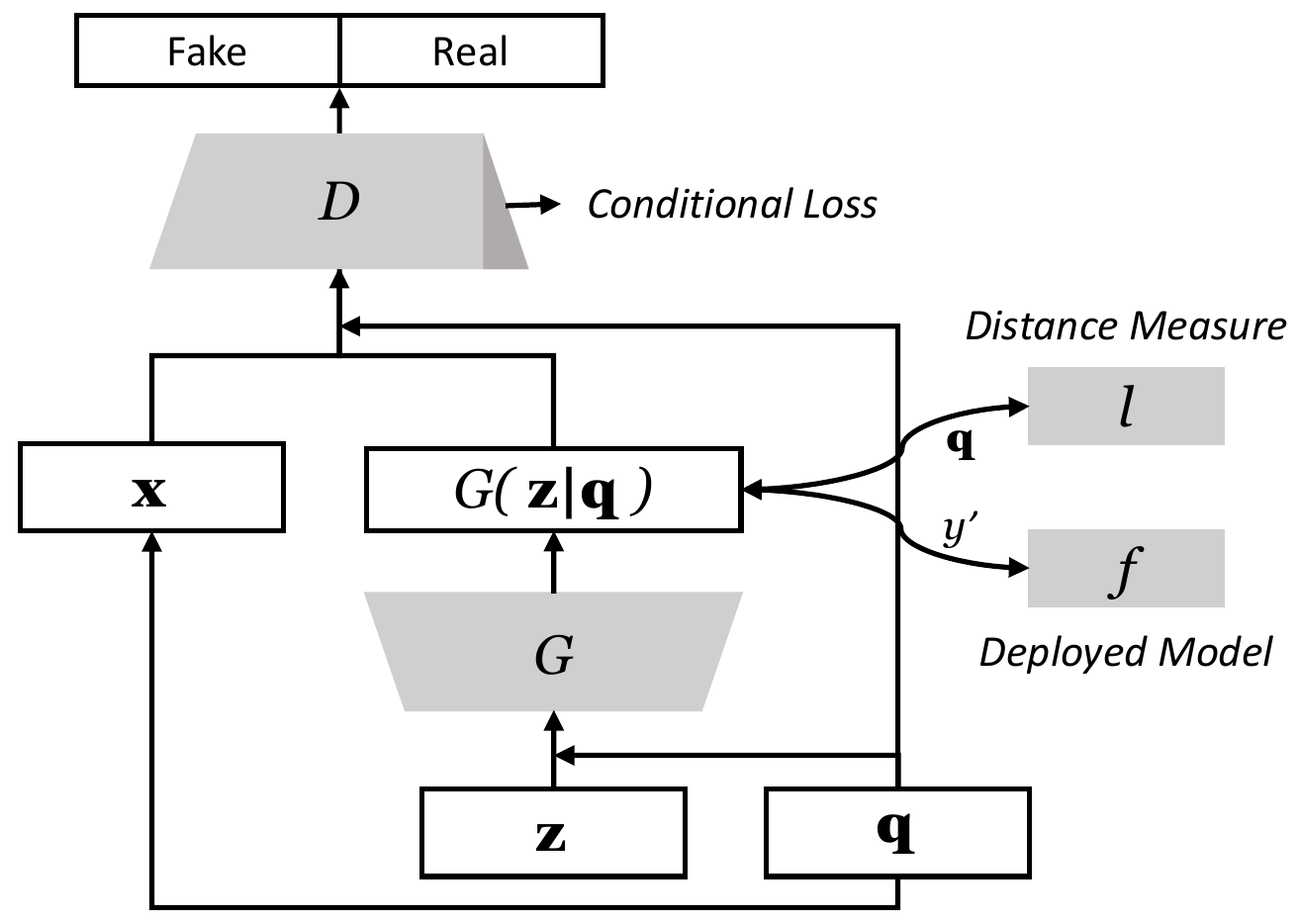}
\vspace{-0.25cm}
\caption{The overall framework design for the proposed counterfactual synthesizer through conditional modeling.} 
\label{fig:cfsyn}
\end{figure}

Designing a model-based synthesizer typically involves how to build an effective generative model for counterfactuals. According to the previous analysis, we know that the key lies in the approximation of hypothetical distribution $\mathcal{C}_{\mathbf{x},\mathbf{y}|\mathbf{q}}$, which can be formulated as a conditional modeling problem. To this end, we propose a conditional generative framework based on CGAN in this paper, specifically designing for counterfactual explanation. The overall architecture is illustrated by Fig.~\ref{fig:cfsyn}. 

In the designed framework, $D$ represents the discriminator module, and $G$ denotes the generator module. Similar as CGAN, $D$ and $G$ are jointly trained as adversaries to each other, aiming to achieve a min-max game. The major difference with CGAN comes from how we prepare the conditional vectors for the framework training. Here, instead of simply employing label information, we use the query as conditions for counterfactual generation. Besides, to guarantee the quality of generated counterfactual samples, we further incorporate a distance measure $l$ and the deployed model $f$ to regularize the training of $G$. Throughout this process, we aim to obtain the counterfactuals which are similar to the query $\mathbf{q}$ and have the preferred output decision $y^{\prime}$. Specifically, the training objective of this counterfactual min-max game can be indicated as:
\begin{equation}\label{eq_mcsobj}
    \min_{G} \max_{D}  \!\!\!\!\!\! \underset{\substack{\quad \mathbf{x}\sim\mathcal{P}_{\mathbf{x}} \\ \quad\mathbf{q}\sim\mathcal{P}_{\mathbf{q}}}}{\mathbb{E}} \!\!\!\!\!\! \log D(\mathbf{x}|\mathbf{q}) \ + \!\!\!\!\!\!
    \underset{\substack{\quad\mathbf{z}\sim\mathcal{P}_{\mathbf{z}} \\ \quad\mathbf{q}\sim\mathcal{P}_{\mathbf{q}}}}{\mathbb{E}} \!\!\!\!\!\! \log \! \left[1\!-\!D(G(\mathbf{z}|\mathbf{q}))\right] \, + \, L^{cf} \!\! \left(G(\mathbf{z}|\mathbf{q})\right),
\end{equation}
where $\mathbf{z}$ represents the noise vector following a distribution $\mathcal{P}_{\mathbf{z}}$. Within Eq.~\ref{eq_mcsobj}, the counterfactual loss term $L^{cf}$ can be further expressed as follows:
\begin{equation}\label{eq:cf_loss}
    L^{cf}\left(G(\mathbf{z}|\mathbf{q})\right) = L^{ce}\left(f(G(\mathbf{z}|\mathbf{q})), \ y^{\prime}\right) + l\left(G(\mathbf{z}|\mathbf{q}), \ \mathbf{q}\right),
\end{equation}
in which $L^{ce}$ indicates the cross-entropy loss between the predictions from $f$ and the preferred decision $y^{\prime}$, regarding to the generated samples. Overall, $L^{cf}$ is expected to be minimized for counterfactual reasoning purposes, and it can be treated as an additional regularization term appended with the conventional CGAN training. By utilizing this conditional design, we can then employ the well-trained generator $G$ to synthesize a series of hypothetical distributions parametrized by query $\mathbf{q}$, and further achieve the model-based counterfactual generation through sampling.

\section{Enhancement for Counterfactuals}

With the previous design, we now consider two practical enhancements for model-based counterfactuals. First, we propose a novel training scheme for counterfactual synthesizer based on the umbrella sampling technique. Second, we utilize the model inductive bias to consider the causal dependence among attributes.

\subsection{Effective Synthesizer Training} 

\subsubsection{Query imbalance during training.}

\begin{figure}
\centering
\includegraphics[width=\columnwidth]{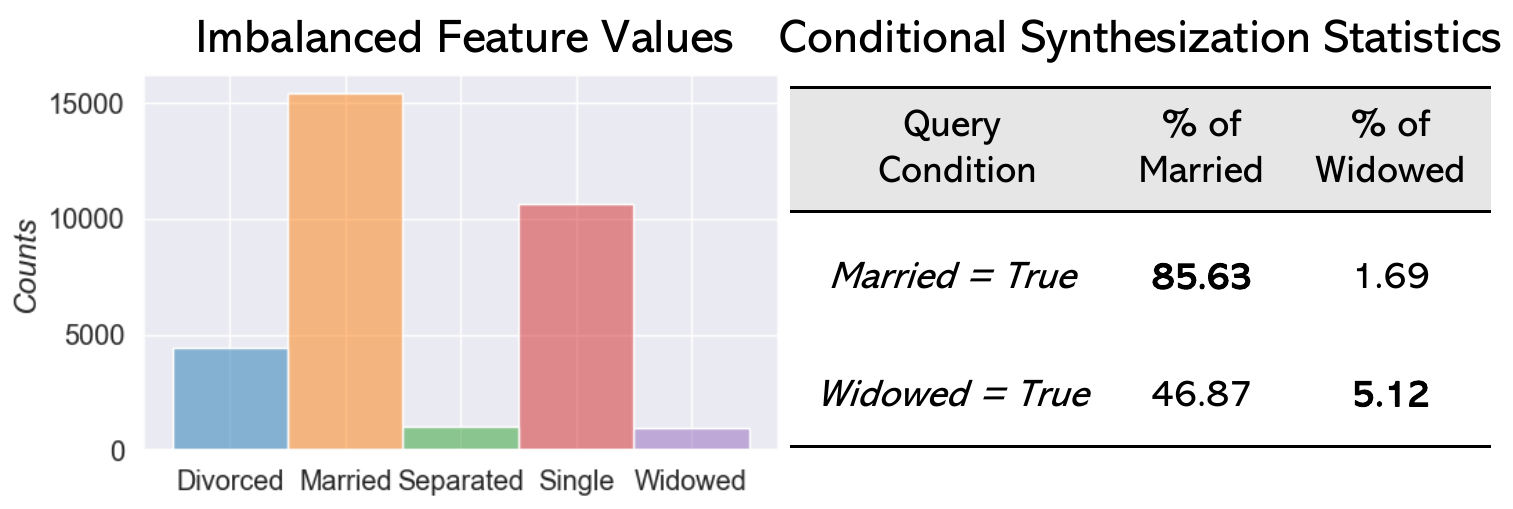}
\vspace{-0.4cm}
\caption{Case results for base synthesizer on conditional performance of the `marital-status' attribute in \textit{Adult} data.} 
\label{fig:case1}
\end{figure}

To effectively train the designed counterfactual synthesizer shown in Fig.~\ref{fig:cfsyn}, we need to let $G$ well capture the conditions indicated by $\mathbf{q}$. However, this may not be as straightforward as in conventional CGAN, since $\mathcal{P}_{\mathbf{q}}$ is typically imbalanced among different attribute values. Such query imbalance results in the fact that the hypothetical distributions conditioned on those rare values cannot be effectively approximated, due to the limited number of instances. To illustrate the point, we show some case results of the base synthesizer on \textit{Adult} dataset\footnote{http://archive.ics.uci.edu/ml/datasets/Adult} in Fig.~\ref{fig:case1}. Here, we assume that $\mathbf{q}$ and $\mathbf{x}$ share a same prior distribution (i.e., $\mathcal{P}_{\mathbf{q}}=\mathcal{P}_{\mathbf{x}}$), because queries are usually collected from similar data sources in most real-world scenarios. According to the results, we note that the imbalanced values of `marital-status' attribute\footnote{We merge \texttt{Married-civ-spouse}, \texttt{Married-spouse-absent}, \texttt{Married-AF-spouse} in `marital-status' attribute all to value \texttt{Married} for simplicity.} lead to significantly different conditional performance on synthesis. When conditioned on those majority values (e.g., \texttt{Married}), the synthesizer can reasonably capture the corresponding hypothetical distribution. In contrast, the synthesizer fails when conditioned on minority values (e.g., \texttt{Widowed}), and its conditional performance is bad. Thus, to better train the designed counterfactual synthesizer, we need to prepare a proper set of training queries, which contains sufficient samples with attribute values in the tails of prior $\mathcal{P}_{\mathbf{q}}$. 

\subsubsection{Umbrella sampling for rare instances.}  

To properly reflect the influence of rare values, we need some enhanced sampling strategies for training, instead of the simple random way. An intuitive way is to relatively increase the probability mass for rare values. In work~\cite{xu2019modeling}, the authors used the frequency logarithm to curve the probability mass, aiming to make the sampling process have higher chances in obtaining those "tail" values. However, such mass curves may not be well suited for our training scenario, because it distorts the prior $\mathcal{P}_{\mathbf{q}}$ and further leads to an unfaithful hypothetical distribution $\mathcal{C}_{\mathbf{x},\mathbf{y}|\mathbf{q}}$. Considering this, we novelly apply the umbrella sampling technique here to enhance the synthesizer training, which was originally used in computational physics and chemistry for molecular simulation~\cite{kastner2011umbrella}. Umbrella sampling recasts the whole sampling process into several unique samplings over umbrella-biased distributions in a weighted-sum manner, where the added artificial umbrellas are expected to cover the full value domain with overlaps. By calculating the weight of each biased distribution, we can then reconstruct the original distribution and conduct evaluations with the umbrella samples obtained. Specifically for counterfactual synthesizer training, we can thus guarantee the sufficient number of queries with balanced values by sampling under different umbrella biases. In particular, the corresponding weight of each biased distribution for query preparation can be calculated as below. 

\begin{theorem}\label{theo1}
    Consider the sampling process $\mathbf{q} \sim \mathcal{P}_{\mathbf{q}}$ for counterfactual synthesizer training. Let $\mathbf{w}=[w_{1}, \cdots, w_{N}]$ denote the weight vector for $N$ umbrella-biased distributions~\cite{kastner2011umbrella}, where $w_{i}$ indicates the normalized weight of the $i$-th biased distribution $\mathcal{P}_{\mathbf{q}}^{i}$, and $\mathbf{u}=[u_{1}, \cdots, u_{N}]$ denote the profile of the added artificial umbrellas. Then, the optimal $\mathbf{w}$ can be derived by solving the equation $\mathbf{w} \mathbf{M} = \mathbf{w}$, where $\mathbf{M}=\mathbf{M}(\mathbf{w})$ represents the overlap matrix defined as:
    \begin{equation}\label{ovlp_mx}
        M_{ij} = \left<\frac{u_{j}/w_{i}}{\sum_{k=1}^{N}u_{k}/w_{k}}\right>_{\mathcal{P}_{\mathbf{q}}^{i}}.
    \end{equation}
\end{theorem}

\noindent Here, the operation $\left<\cdot\right>_{\mathcal{P}}$ indicates the average over distribution $\mathcal{P}$. The proof for deriving $\mathbf{w}$ is shown in Appendix~\ref{proof1}.

\subsection{Causal Dependent Generation}

\subsubsection{Causality for generated counterfactuals.}

\begin{figure}
\centering
\includegraphics[width=0.95\columnwidth]{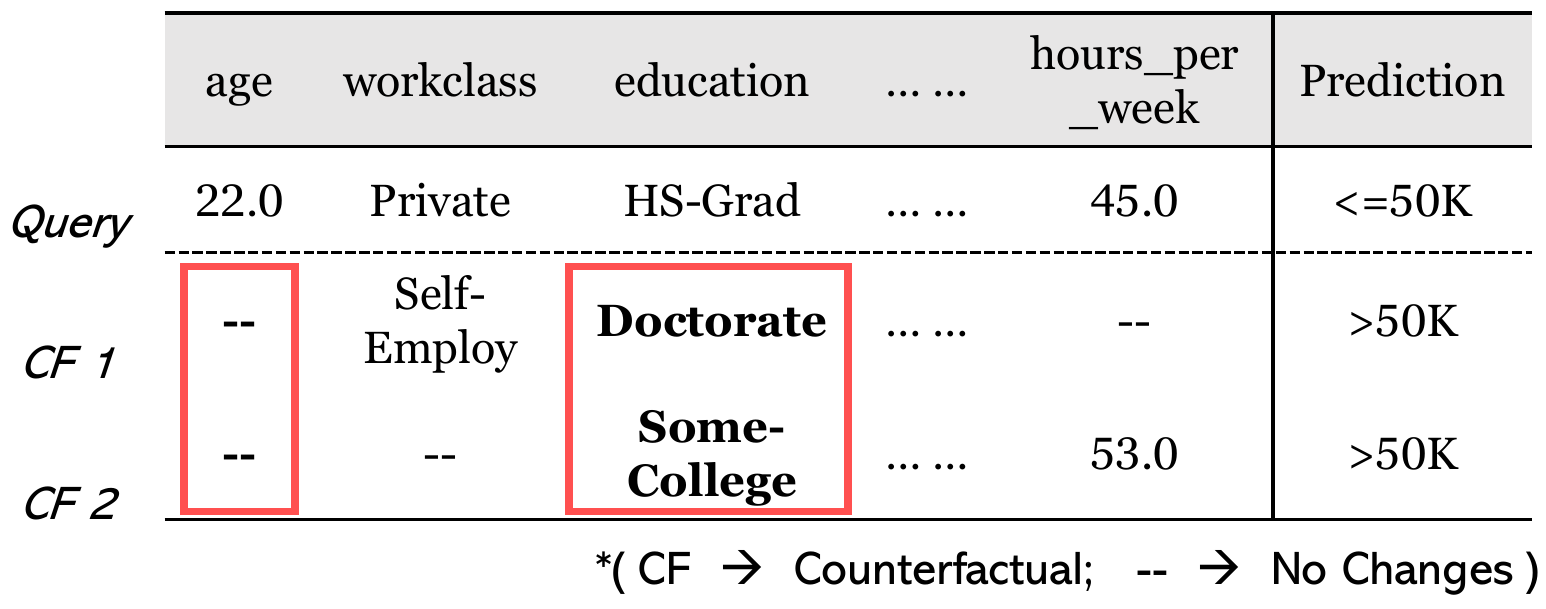}
\vspace{-0.3cm}
\caption{Case study on counterfactual samples (selected) generated by DiCE algorithm in \textit{Adult} data.} 
\label{fig:case2}
\end{figure} 

To make the generated counterfactuals have better feasibility on reasoning, it is also preferred to consider the causality among different input attributes. We here use another case study on the \textit{Adult} data, shown in Fig.~\ref{fig:case2}, to illustrate the point. In particular, we employ an existing counterfactual explanation method, DiCE~\cite{mothilal2020explaining}, to generate related counterfactual samples. The results show in Fig.~\ref{fig:case2} that DiCE is not able to reflect the causality between the attribute `education' and `age', since it suggests to improve the education level without changing the age for flipping model decisions. In real-world settings, such counterfactuals are usually considered infeasible, because input attributes cannot be altered independently in counterfactuals. Specifically in this example, education level typically improves with age, so feasible counterfactuals should indicate such causality for human reasoning. We now enhance our proposed MCS by showing how to incorporate such causal dependencies among attributes.

\subsubsection{Model inductive bias for causal dependence.}

\begin{figure}
\centering
\includegraphics[width=0.95\columnwidth]{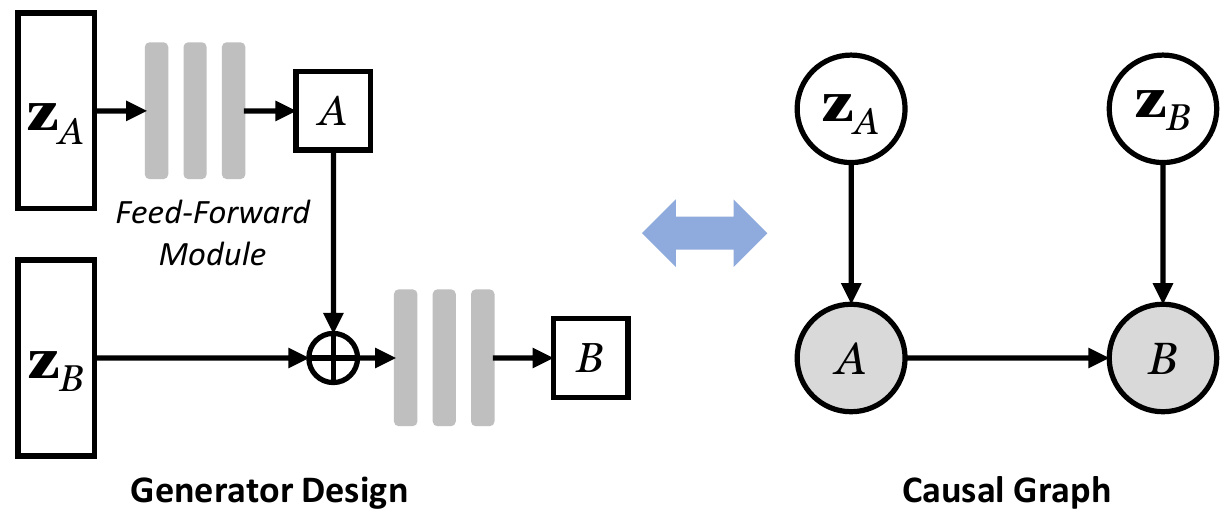}
\caption{Example design of $G$ for incorporating the causation: $A \rightarrow B$, as well as the represented causal graph.}  
\label{fig:se_mib}
\end{figure}

In contrast with the algorithm-based counterfactuals, model-based counterfactuals provide new possibilities to incorporate domain-specific causal dependence for explanation, instead of simply adding extra regularization or constraints for generation~\cite{mahajan2019preserving}. In particular, we propose to utilize the inductive bias of $G$ to encode relevant causal dependence, where the architecture of $G$ is intentionally designed to mimic the structural equations~\cite{hall2007structural} of corresponding causation. Fig.~\ref{fig:se_mib} shows an example of designing $G$ for the causal relationship $A \rightarrow B$, where $A$ indicates the cause and $B$ denotes the effect. Essentially, by purposely manipulating the generator architecture, we encode the causal structure as inductive biases in $G$, so as to achieve the causal dependent generation for counterfactuals. We now state a theorem proving the correctness of this approach from the perspective of causality identification, where the corresponding considered causal dependence is shown to be existed from the generated counterfactual samples. 

\begin{theorem}\label{theo2}
    Let $\mathcal{D}=(\mathcal{A}, \mathcal{E})$ represent a causal graph with vertex set $\mathcal{A}=\{A_{1}, A_{2}, \cdots, A_{S}\}$ being attributes used for counterfactual generation, and edge set $\mathcal{E}$ consisting of directed edges from causes to effects. Then, the generator $G$ can be designed with $S$ feed-forward modules $G=[F_{1},F_{2},\cdots,F_{S}]$, mimicking the corresponding structural equations of $\mathcal{D}$, such that $A_{s}$ can be generated by
    \begin{equation}
        A_{s}=F_{s}\left(\{A_{p}\}_{p \in \mathrm{Pa}_{s}^{\mathcal{D}}} \ ,\mathbf{z}_{A_{s}}\right), \quad \forall s \in [1,\cdots,S],
    \end{equation}
    where $\mathrm{Pa}_{s}^{\mathcal{D}}$ denotes the set of parent attributes (i.e., causes) for $A_{s}$ in graph $\mathcal{D}$. As a result, the counterfactual samples generated by $G$ are further guaranteed to have the incorporated causality, which can be identified from the observational perspective.
\end{theorem}

\noindent The relevant discussion \& proof are shown in Appendix~\ref{proof2}.

\section{Implementation}

In this section, we briefly introduce the employed practical techniques when implementing the proposed MCS framework. We also show the overall pipeline of MCS for counterfactual generation.

\textbf{Data Representation.} When implementing MCS, we employ different modeling techniques for different types of data attributes. We use Gaussian mixture models~\cite{bishop2006pattern} to encode continuous attributes, and normalize the values according to the selected mixture mode. We represent discrete attributes directly using one-hot encoding. Furthermore, one specific data instance $\mathbf{r}_{ins}$ can be then represented as
\begin{equation}\label{dm}
    \mathbf{r}_{ins}=\mathbf{c}_{1}^{ins} \oplus \cdots \oplus \mathbf{c}_{N_{c}}^{ins} \oplus \mathbf{d}_{1}^{ins} \oplus \cdots \oplus \mathbf{d}_{N_{d}}^{ins}, 
\end{equation}
where $N_{c}$ and $N_{d}$ indicate the number of continuous and discrete attributes in the data respectively. Also, we represent the query $\mathbf{q}$ by value masking, thus giving humans control over the semantics of the counterfactuals as appropriate for any particular use case. 

\textbf{Umbrella Sampling for Discrete Attributes.} Conventional umbrella sampling technique only applies to continuous distributions. For discrete attributes, we use the Gumbel-Softmax~\cite{jang2016categorical} method to relax the categorical distribution into a continuous one. This reparameterization trick is proved to be faithful and effective in many cases with appropriate temperature $\tau$, which controls the trade-off of the distribution relaxation. When $\tau \rightarrow 0$, the relaxed distribution becomes into the original discrete one. When $\tau \rightarrow \infty$, it gradually converges into a uniform distribution. In our proposed MCS framework, we conduct the implementation with a fixed temperature as $\tau=0.5$. 

\textbf{Overall MCS Pipeline.} To clearly show the procedures of building MCS for counterfactual explanation, we give an overview of the pipeline, illustrated by Algorithm~\ref{algo}. Compared with the existing algorithm-based counterfactuals, the proposed MCS is much more efficient during the interpretation phase, since it avoids the iterative perturbation step regarding to each input $\mathbf{q}$. In exchange, MCS pushes the computational complexity to the setup and training phase, which largely depends on the data scale as well as the value space we focus on.

\begin{algorithm}[t]

\# \textit{Setup Phase}

- Prepare $f$ for explanation, and select $l$ for measurement; \\
- Design the generator $G$ with domain $\mathcal{D}$ based on Theo.~\ref{theo2};

\# \textit{Training Phase}

- Data Modeling with Eq.~\ref{dm}; \\
- \For{training batch $k$}{
    Utilize umbrella sampling to prepare a set of queries $\mathbf{q}_{k}$; \\
    Weigh batch $k$ with $\mathbf{w}$ over $\mathbf{q}_{k}$ based on Theo.~\ref{theo1}; \\
    Update $G$ and $D$ in the min-max game of Eq.~\ref{eq_mcsobj}; 
    }
    
\# \textit{Interpretation Phase}

- Feed the user query $\mathbf{q}$ to $G$ for counterfactual generation.

\caption{Building Pipeline of MCS for Interpretation}
\label{algo}
\end{algorithm}

\section{Experiments} 

In this section, we empirically evaluate the proposed MCS on both synthetic and real-world datasets from several different aspects, and aim to answer the following key research questions: 
\begin{itemize}[leftmargin=*]

    \item How effective and efficient is MCS in generating counterfactual samples, compared with existing algorithm-based methods? 
    
    \item How well does MCS model the original observational distribution as well as the conditional counterfactual distribution, with the umbrella sampling technique? 
    
    \item Can we identify the incorporated domain-specific causal dependence from the counterfactual samples generated by MCS?

\end{itemize}

\subsection{Evaluation on Counterfactual Generation}

In this part, we evaluate the effectiveness and efficiency of MCS, comparing with existing algorithm-based counterfactual methods. 

\subsubsection{Experimental settings.} 

\begin{table}[t] 
\caption{Data statistics in experiments.}
\vspace{-0.4cm}
\centering
\setlength{\tabcolsep}{3.9pt} 
\begin{normalsize}
\begin{tabular}[width=1cm]{cccccc}
\toprule
Dataset        & \#Row      & \#Col    & Attribute Type  \\
\midrule
Syn\_Moons       & $500$       & $3$    & Continuous      \\  
Syn\_Circles     & $500$       & $3$    & Continuous      \\  
Adult            & $48,842$     & $9$         & Continuous \& Categorical   \\
Home\_Credit            & $344,971$        & $39$     & Continuous \& Categorical   \\
\bottomrule
\end{tabular}
\end{normalsize}
\label{tab:data}
\end{table}

We consider two synthetic datasets and two real-world datasets for counterfactual generation evaluation. Specifically, the statistics of the datasets are shown in Tab.~\ref{tab:data}. 
\begin{itemize}[leftmargin=*]
\item \textbf{\textit{Synthetic}}\footnote{https://scikit-learn.org/stable/modules/classes.html\#module-sklearn.datasets}: We synthesize two datasets for classification purpose, i.e., ``Syn\_Moons'' and ``Syn\_Circles'', with different separation boundaries. To facilitate visualization, the synthetic data only contains two continuous attributes. These two synthetic datasets are mainly used to evaluate the MCS effectiveness. 

\item \textbf{\textit{Adult}}: This is a real-world benchmark dataset for income prediction, where each instance is labelled as ``>50K'' or ``<=50K''. In the experiments, the counterfactuals on this data aim to help understand how to flip model decisions from ``<=50K'' to ``>50K''. To facilitate our task, we only consider a subset of the attributes. 

\item \textbf{\textit{Home\_Credit}}\footnote{https://www.kaggle.com/c/home-credit-default-risk/data}: This is a larger real-world dataset for client risk assessment, where the goal is to predict clients' repayment abilities of given loans. The counterfactuals here are to help reason how to make improvements for risky clients to become non-risky. We drop some unimportant attributes in experiments. 
\end{itemize}

\noindent As for the deployed classifier, we prepare three different $f$ for counterfactual generation evaluation as below. Those classifiers are all trained with $80\%$ of the data, and tested with the rest $20\%$. 
\begin{itemize}[leftmargin=*]
\item \textbf{\textit{RBF SVM}}: This is a pre-trained support vector machine (SVM) with the RBF kernel, where a squared $l_2$ penalty is applied. Related SVM hyperparameter $\gamma$ is set to $2$, and $C$ is set to $1$. 

\item \textbf{\textit{Random Forest (RF)}}: This is a pre-trained tree-based RF classifier with $10$ estimators. The maximum depth of each tree is set as $5$. We use the Gini impurity as our splitting criterion. 

\item \textbf{\textit{Neural Net (MLP)}}: This is a pre-trained ReLU neural classifier with multi-layer perceptron (MLP). Here, we have $1$ hidden layer and $100$ hidden units. The relevant $l_2$ regularization coefficient is set to $1$, and the maximum iteration number is set as $1000$. 
\end{itemize}

\noindent Furthermore, we select four recent algorithm-based counterfactual explanation methods as our baselines for comparison. These methods are all set up with their default settings. 
\begin{itemize}[leftmargin=*]
\item \textbf{\textit{DiCE}}~\cite{mothilal2020explaining}: This method generates diverse counterfactual samples by providing feature-perturbed versions of the query, where the perturbations are derived by iterative optimization. 

\item \textbf{\textit{C-CHVAE}}~\cite{pawelczyk2020learning}: This method utilizes a pre-trained variational auto-encoder (VAE) to transform the data space into a latent embedding space, and then perturbs the latent representation of the query for counterfactual generation. 

\item \textbf{\textit{CADEX}}~\cite{moore2019explaining}: This method employs the gradient-based scheme to perturb the query for flipping outcomes, which is an application of adversarial attack methods for counterfactual generation. 

\item \textbf{\textit{CLEAR}}~\cite{white2019measurable}: This method uses the concept of $b$-perturbation to construct potential counterfactuals through local regression, where the corresponding fidelity error is minimized iteratively.  
\end{itemize}

\subsubsection{Counterfactual generation effectiveness.} 

\begin{figure}
\centering
\includegraphics[width=\columnwidth]{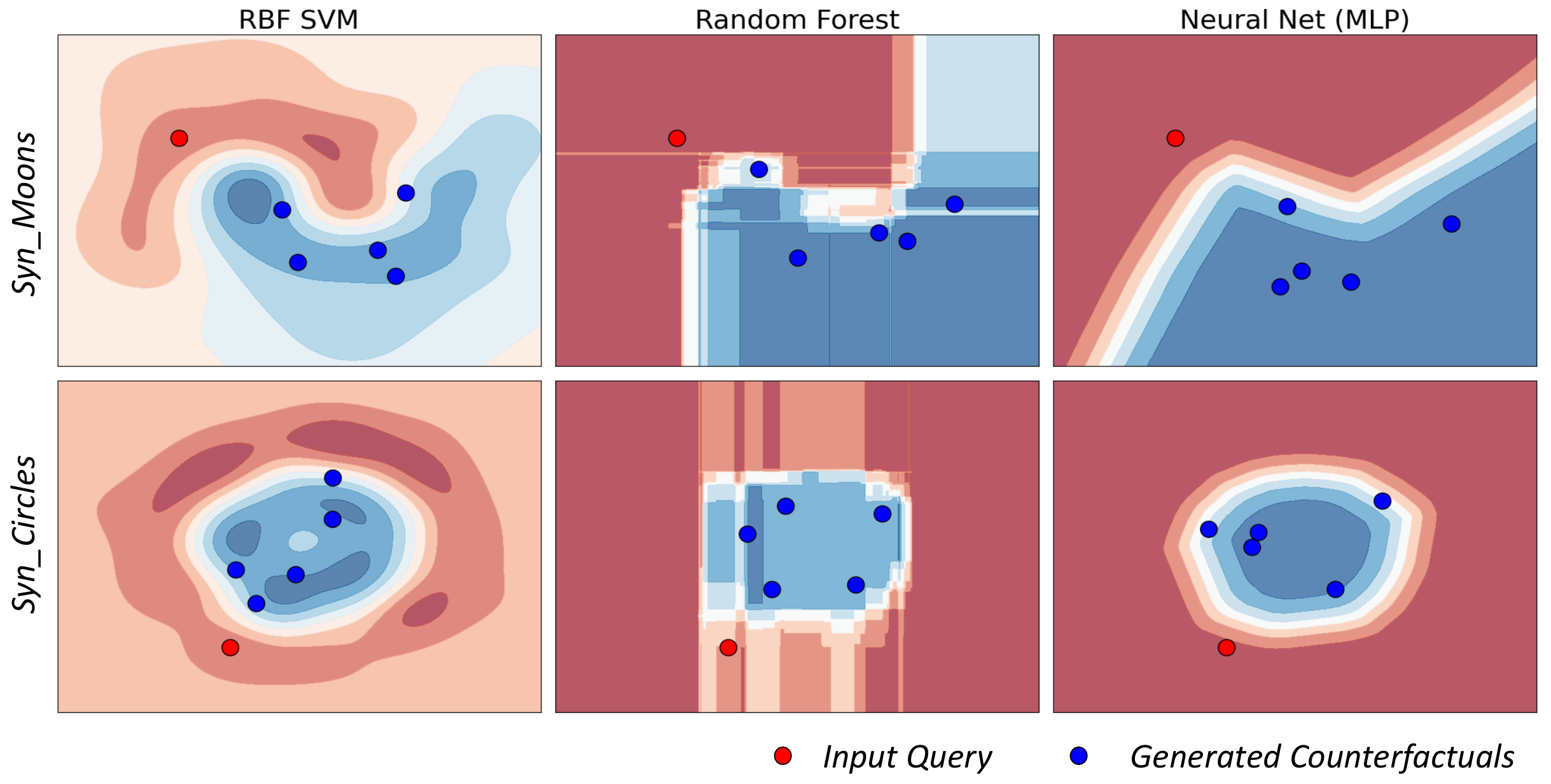}
\vspace{-0.65cm} 
\caption{Visualization of the generated counterfactual samples by the proposed MCS on two synthetic datasets.} 
\label{fig:cf_gen_bound}
\end{figure} 

\begin{figure*}[t] 
\centering
\includegraphics[width=\textwidth]{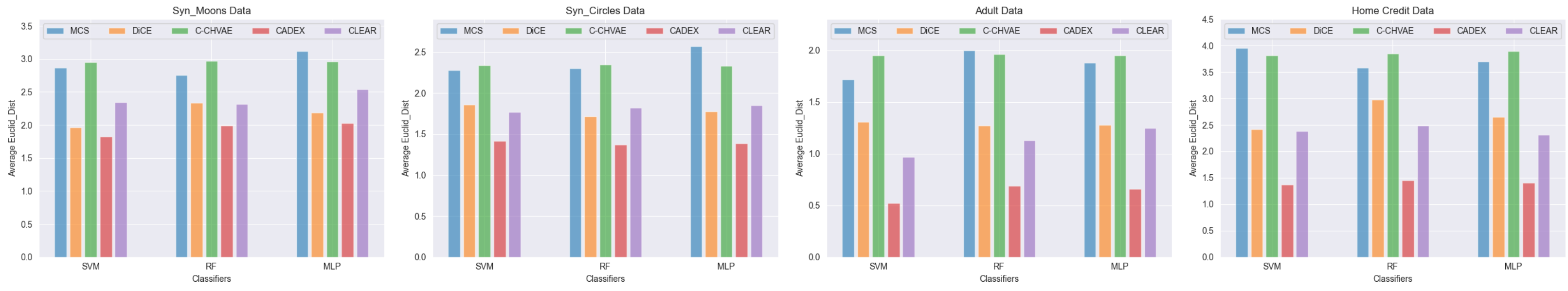}
\vspace{-0.7cm}
\caption{The average Euclidean distance comparison between the query and the generated counterfactuals.}
\label{fig:distcomp}
\end{figure*} 

To demonstrate the effectiveness of MCS, we visualize the generated counterfactuals on two synthetic datasets, illustrated by Fig.~\ref{fig:cf_gen_bound}. In this set of experiments, we select a fixed query $\mathbf{q}_{0}$ for each synthetic dataset with negative model decision (i.e., predicted as `$-1$'), and randomly show $5$ samples generated by MCS. Fig.~\ref{fig:cf_gen_bound} shows that all generated samples successfully flip the query prediction from the negative `$-1$' to the positive `$1$', across different classifiers $f$. Thus, it is noted that MCS is capable of synthesizing valid counterfactuals for prediction reasoning. To further evaluate the counterfactuals generated by MCS, we employ the \emph{average Euclidean distance} as the metric, indicated by Eq.~\ref{eq:avg_dis}, to reflect how close of the generated samples regarding to the input query: 
\begin{equation}
\slfrac{\delta^{\mathrm{avg}} = \sum\nolimits_{\mathbf{e} \in \mathcal{S}^{cf}}\mathrm{Euclid\_Dist}(\mathbf{q}_{0}, \mathbf{e})}{\left|\mathcal{S}^{cf}\right|},
\label{eq:avg_dis}
\end{equation}
where $\mathcal{S}^{cf}$ denotes the set of the generated counterfactuals. In our experiments, we set $|\mathcal{S}^{cf}|=20$, and compare $\delta^{\mathrm{avg}}$ among different counterfactual methods under different $f$ over different datasets. Fig.~\ref{fig:distcomp} shows that counterfactuals generated by MCS are generally farther to the queries, compared with those generated by baselines (except C-CHVAE). This observation suggests that algorithm-based methods are good at finding those ``nearest'' counterfactuals through iterative perturbations, while MCS mainly focus on the counterfactual distribution approximation instead of sample searching. For C-CHVAE, it is designed to conduct perturbations in the latent space, and hence its generated counterfactuals are also observed to be farther than those from DiCE, CADEX and CLEAR in the data space. C-CHVAE can be treated as a data density approximator, thus the corresponding $\delta^{\mathrm{avg}}$ are almost the same over different $f$ within the same dataset. Overall, we know that MCS can effectively generate valid counterfactuals for reasoning, but its synthesized samples may not be as close to the query as those generated from existing algorithm-based methods (i.e., DiCE, CADEX, CLEAR), which becomes more significant with the increase of the data scale.

\subsubsection{Counterfactual generation efficiency.} 

\begin{figure}
\centering
\includegraphics[width=\columnwidth]{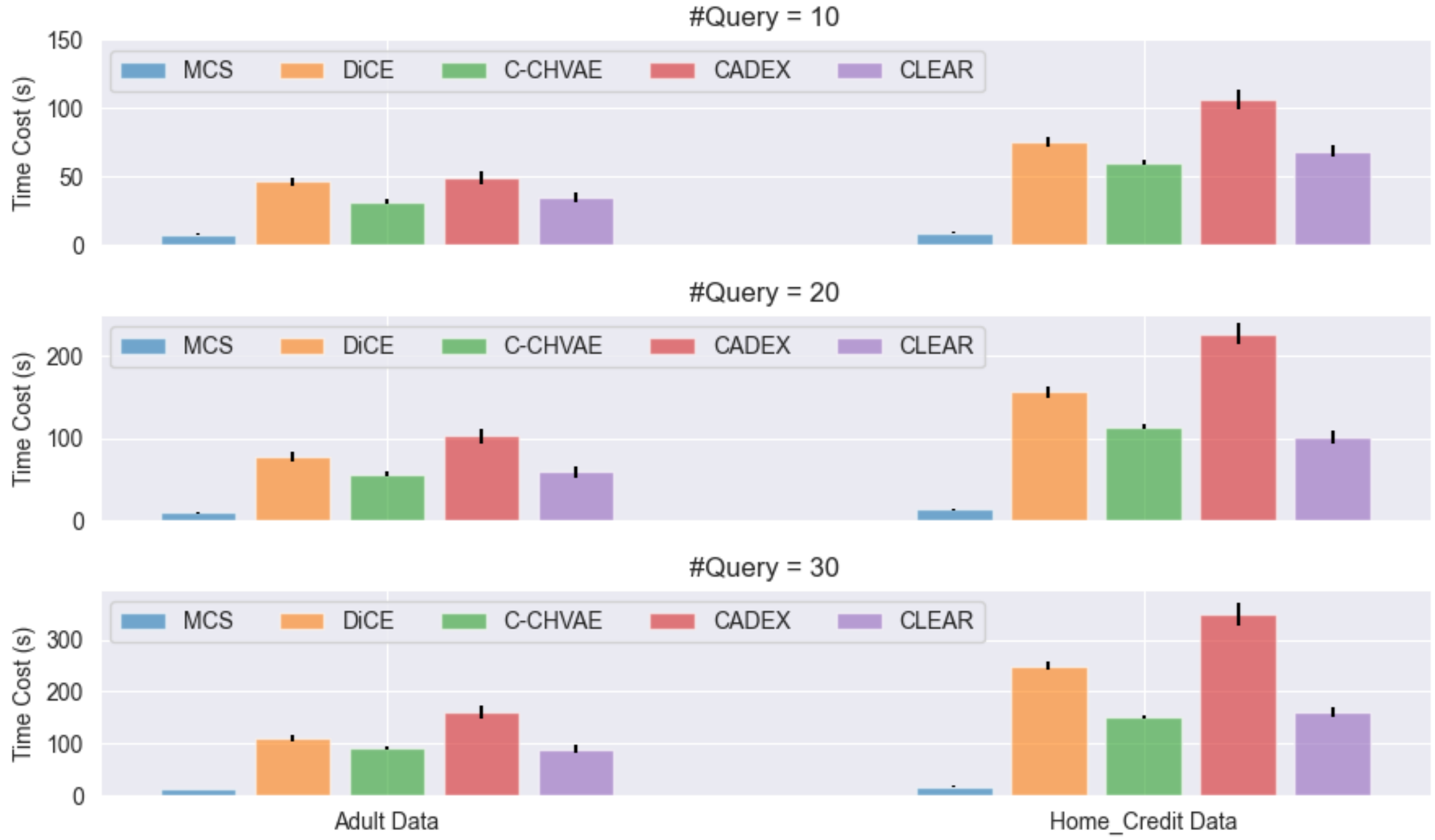}
\vspace{-0.65cm} 
\caption{The time efficiency comparison with MLP.} 
\label{fig:cf_exp_time}
\end{figure} 

To evaluate the efficiency of different counterfactual explanation methods, we compare the time cost of sample generation process regarding to multiple queries. In this set of experiments, we only consider the \textit{Adult} and \textit{Home\_Credit} dataset for illustration, with a MLP classifier $f$, and test the counterfactual generation under $10$, $20$ and $30$ queries. The relevant results are reported by averaging $5$ runs over different sets of input. Fig.~\ref{fig:cf_exp_time} shows that MCS consumes significantly less time for counterfactual generation, and such merit over algorithm-based methods is more remarkable with a larger scale of input. Besides, we can also observe that MCS is much more stable (i.e., lower standard deviation) among different queries, and its unit time cost for each query almost keeps identical despite the specific inputs.

\subsection{Evaluation on Distribution Modeling}

In this part, we evaluate the modeling performance of MCS aided by umbrella sampling technique. Overall, we aim to demonstrate the faithfulness of the generated counterfactuals. 

\subsubsection{Experimental settings}
We only consider the two real-world datasets (i.e., \textit{Adult} and \textit{Home\_Credit} data) for this part of evaluation. For the artificial umbrellas used during MCS training, we respectively set $N=8$ and $N=32$ for training query sampling in the \textit{Adult} and \textit{Home\_Credit} dataset. Detailed hyper-parameters, as well as the related influence studies, of the umbrella sampling process are introduced in Appendix~\ref{appendix_us}.

\subsubsection{Observational distribution modeling.} 

\begin{table}[t]
\caption{Average F-score of testing on learning efficacy.}
\vspace{-0.39cm}
\centering
\label{tab:obsmodel}
\begin{tabular}{c|cc|cc|cc}
\toprule
\multirow{2}{*}{Dataset} & \multicolumn{2}{c|}{$\mathcal{F}_{o}$} & \multicolumn{2}{c|}{$\mathcal{F}_{t}$} & \multicolumn{2}{c}{$\mathcal{F}_{p}$} \\ \cline{2-7} 
                         & RF        & MLP       & RF        & MLP       & RF       & MLP       \\ \hline
Adult                    & 0.616          & 0.435          & 0.593          & 0.429          & 0.586         & 0.431          \\
Home\_Credit             & 0.602          & 0.386          & 0.421          &  0.307         & 0.418         & 0.306          \\ 
\bottomrule
\end{tabular}
\end{table}

To evaluate the MCS modeling on observational data distribution, we use the metric \textit{model compatibility} in~\cite{park2018data} focusing on learning efficacy. The intuition behind this metric is that, models trained on data with similar distributions should have similar test performances. Furthermore, to conduct such evaluation, we prepare three different sets of classifiers (i.e. $\mathcal{F}_{o}$, $\mathcal{F}_{t}$ and $\mathcal{F}_{p}$) for test comparison, which are respectively trained on the original data, table-GAN~\cite{park2018data} synthesized data and MCS synthesized data. Here, we employ table-GAN as a baseline synthesizer, since it has been proven to be an effective way to create hypothetical samples which follow a particular observational distribution. As for MCS, we regard the label $\mathbf{y}$ as an additional attribute for synthesis in this part, and remove $L^{cf}$ in Eq.~\ref{eq_mcsobj} during training. The final synthesized data is then generated by $G$ in an unconstrained manner with $\mathbf{q}$ set to \textit{None} (all zero). In experiments, we consider two types of classifiers (RF \& MLP), and report the average F-score over five rounds. The results in Tab.~\ref{tab:obsmodel} show that classifiers in $\mathcal{F}_{p}$ have competitive test performances with those in $\mathcal{F}_{t}$, and thus demonstrates that MCS can reasonably model the observational distribution with its synthesized data samples.

\subsubsection{Counterfactual distribution modeling.} 

\begin{figure}
\centering
\includegraphics[width=\columnwidth]{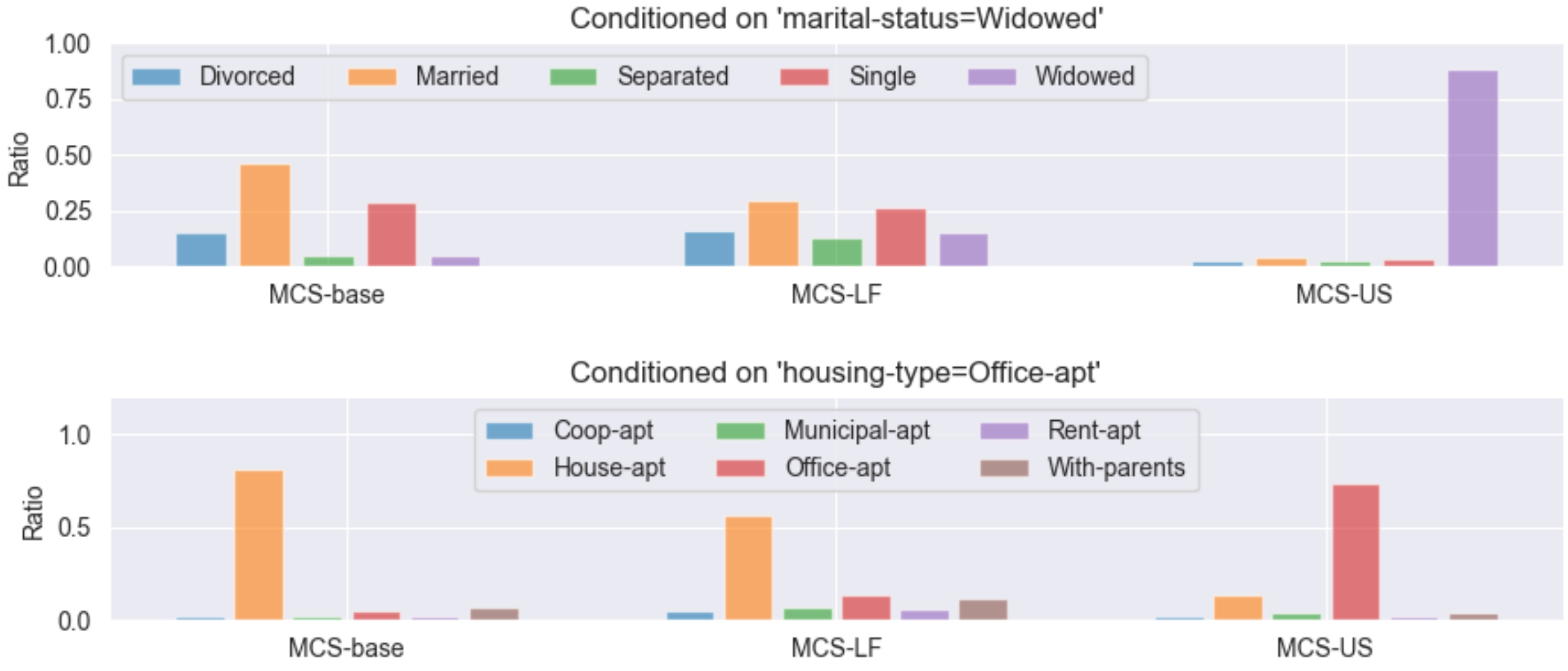}
\vspace{-0.5cm} 
\caption{The counterfactual distributions derived by MCS.} 
\label{fig:cf_distribution}
\end{figure}

Evaluating MCS in modeling counterfactual distribution is essentially to assess the conditional generation performance under particular human priors. When users request hypothetical samples for reasoning, a good synthesizer should have a reasonable conditional performance in generating those in-need samples which are consistent with human priors. In the experiments herein, we select the `marital-status' and `housing-type' attributes in \textit{Adult} and \textit{Home\_Credit} data for illustration, considering assumptive priors on rare values for testing. In particular, we evaluate with priors on `marital-status=\texttt{Widowed}' and `housing-type=\texttt{Office-apt}'. Originally, \texttt{Widowed} takes around $3\%$, and \texttt{Office-apt} takes around $1\%$. The derived counterfactual distributions from MCS of `marital-status' and `housing-type' are illustrated by Fig.~\ref{fig:cf_distribution}. Here, \textit{MCS-base} indicates the synthesizer trained directly with Eq.~\ref{eq_mcsobj}, \textit{MCS-LF} represents the synthesizer trained with logarithm frequency curve~\cite{xu2019modeling}, and \textit{MCS-US} denotes the synthesizer trained with the umbrella sampling technique. Based on the results, it is noted that, aided with umbrella sampling, our proposed MCS can effectively approximate the hypothetical distribution presumed by human priors, and generate in-need counterfactuals to facilitate the reasoning process. 

\subsection{Evaluation on Causal Dependence} 

In this part, we evaluate the causal dependence over the samples from MCS, and test the causation reflected by model inductive bias. 

\subsubsection{Experimental settings.} 
We only use \textit{Adult} and \textit{Home\_Credit} data for causal dependence evaluation. When designing the generator $G$, we consider the following cause-effect pairs for synthesis:   
\begin{equation}\label{eq:causation}
\begin{split}
Adult &: \ \mathrm{education} \rightarrow \mathrm{age};  \\
Home\_Credit &: \ \mathrm{income\_type} \rightarrow \mathrm{income\_total}.
\end{split}
\end{equation}
To validate such pairwise causality in synthesized data, we employ two different methods that are commonly used as below:
\begin{itemize}[leftmargin=*]
\item \textbf{\textit{ANM}}~\cite{hoyer2008nonlinear}: It is a popular approach for pairwise causality identification, which bases on the data fitness to the additive noise model on one direction and the rejection on the other direction. 

\item \textbf{\textit{CDS}}~\cite{fonollosa2019conditional}: It measures the variance of marginals after conditioning on bins, which indicates statistical features of joint distribution. 
\end{itemize}

\subsubsection{Pairwise causation identification.}  
\begin{figure}
\centering
\includegraphics[width=\columnwidth]{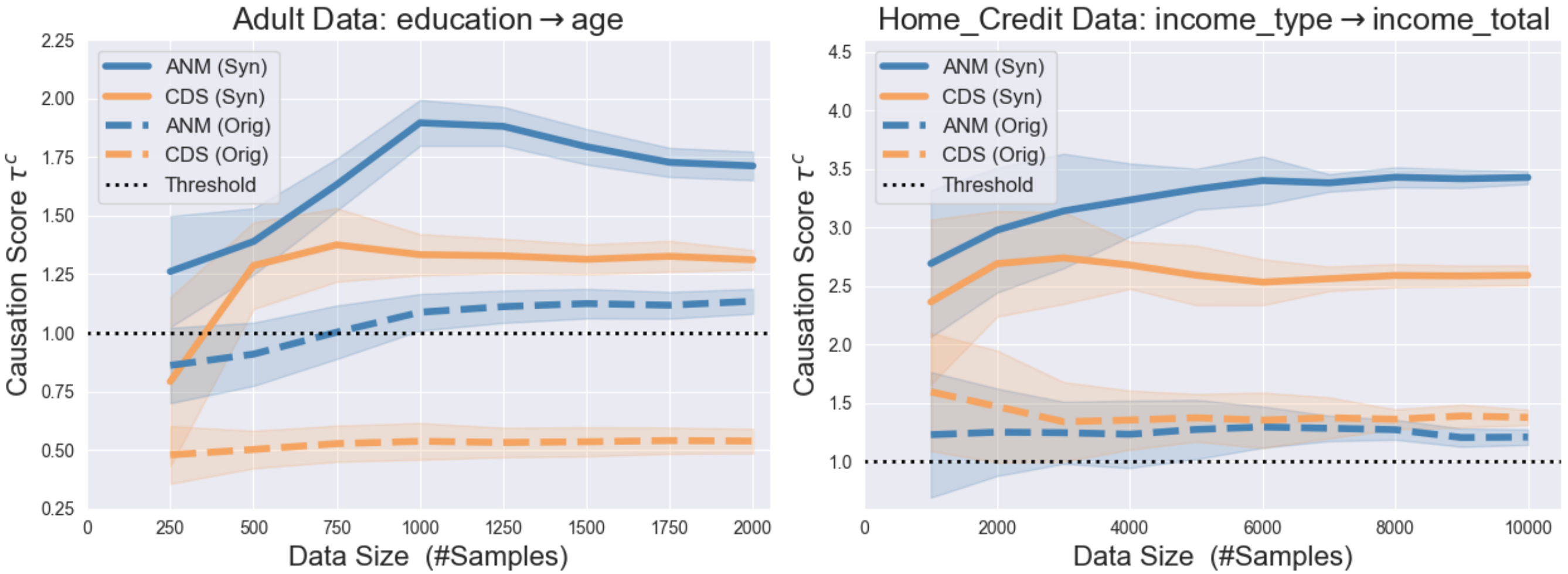}
\vspace{-0.5cm} 
\caption{The scores of considered pairwise causation.} 
\label{fig:cf_cs}
\end{figure}

To conduct the evaluation on pairwise causal dependence, we utilize a simple causation score $\tau^{c}$ to indicate the strength of particular cause-effect pairs. Specifically for $A \rightarrow B$, the causation score can be calculated as follows:
\begin{equation}
    \tau^{c}_{A \rightarrow B} = \tau^{f}_{B \rightarrow A} - \tau^{f}_{A \rightarrow B} \ ,
\end{equation}
where $\tau^{f}$ denotes the data fitness score for a given direction. For different methods, $\tau^{f}$ has different statistical meanings. In our case, $\tau^{f}$ indicates an independence test\footnote{We test the independence using the Hilbert-Schmidt independence criterion (HSIC).} score for ANM, and represents the standard deviation of fitness for CDS. Thus, we know that, the larger the $\tau^{f}$ is, the causation on the given direction is less likely to happen from the observational perspective. In practice~\cite{kalainathan2020causal}, the causation $A \rightarrow B$ is considered to exist when $\tau^{c}_{A \rightarrow B} \geq 1$, and $B \rightarrow A$ when $\tau^{c}_{A \rightarrow B} \leq -1$. When $-1 < \tau^{c}_{A \rightarrow B} < 1$ holds, it usually indicates that there is no obvious causal dependence identified based on statistical analysis. In experiments, we test the causation pairs in Eq.~\ref{eq:causation}, and show the relevant results in Fig.~\ref{fig:cf_cs}. Here, we compare $\tau^{c}$ between the original and synthesized data with both ANM and CDS under different data scale. We observe a stronger causal dependence for the considered pairs in the synthesized data, which demonstrates the effectiveness of MCS on causal dependent generation. With such advantages, humans can easily obtain feasible counterfactuals by properly incorporating relevant domain-specific knowledge into MCS for interpretation.  

\section{Related Work} 

Counterfactuals are one of many interpretation techniques for ML models. In general, according to the format of \textit{explanation carrier} (i.e., how explanations are delivered to humans), related interpretation methods can be categorized as follows.

The first category of interpretation methods uses instance features as indicators to demonstrate which part of the input contributes most to the model predictions. A representative work is \textit{LIME}~\cite{ribeiro2016should}, which uses linear models to approximate the local decision boundary and derive feature importance by perturbing the input sample. Some other methods, utilizing input perturbations for feature importance calculation, can also be found in \textit{Anchors}~\cite{ribeiro2018anchors} and \textit{SHAP}~\cite{lundberg2017unified}. Besides, employing model gradient information for feature attribution is another common methodology under this category. Related examples can be found in \textit{GradCAM}~\cite{selvaraju2017grad} and \textit{Integrated Gradients}~\cite{sundararajan2017axiomatic}. 

The second category uses abstracted concepts as high-level features to indicate the prediction attribution process. One of the earliest work in this category is \textit{TCAV}~\cite{kim2018interpretability}, which trains linear concept classifiers to derive concept representations, and measures the concept importance based on sensitivity analysis. Besides, the authors in~\cite{zhou2018interpretable} decompose the model prediction semantically according to the projection onto concept vectors, and quantify the contributions over a large concept corpus. Inspired by \textit{SHAP}, a corresponding attribution method with human concepts, named \textit{ConceptSHAP}, is proposed in~\cite{yeh2020completeness} to quantify the concept contributions with game theories. Some other follow-up work based on \textit{TCAV} can also be found in~\cite{goyal2019explaining,ghorbani2019towards}. 

The third category uses data samples to deliver relevant explanations to humans. One line of research is to select out prototype or criticism data samples in training set to interpret model behaviors~\cite{kim2016examples,chen2019looks}. Similarly in~\cite{koh2017understanding}, influential training samples to particular model predictions are selected, with the aid of influence functions in measuring sample importance. Furthermore, beyond the real samples in training set, synthesized hypothetical one (e.g., counterfactual sample) is yet another way to interpret predictions for model reasoning. Several representative work along this direction can be found in~\cite{mothilal2020explaining,pawelczyk2020learning,joshi2018xgems}. Our work generally lies in this category of methods for interpreting ML model behaviors.

\section{Conclusions} 

In this paper, a general interpretation framework, named MCS, has been proposed to synthesize model-based counterfactuals for prediction reasoning. By analyzing the focusing counterfactual universe, we first formally defined the problem of model-based counterfactual explanation, and then employed the CGAN structure to train our proposed MCS in an end-to-end manner. To better capture the hypothetical distributions, we novelly applied the umbrella sampling technique to enhance the synthesizer training. Furthermore, we also showed a promising way to incorporate the attribute causal dependence into MCS with model inductive bias, aiming to achieve better feasibility for the derived counterfactuals. Experimental results on both synthetic and real-world data validated several advantages of MCS over other alternatives. Future work extensions may include the model-based counterfactual explorations under more challenging contexts, such as involving high-dimensional data space, time-series sequential nature, or some ethical concerns.

\begin{acks}
The authors would like to thank the anonymous reviewers for their helpful comments. This work is in part supported by NSF IIS-1900990, and NSF IIS-1939716. The views and conclusions contained in this paper are those of the authors and should not be interpreted as representing any funding agencies. This work was partially done when F. Yang was an intern at J.\ P.\ Morgan AI Research.

\textit{Disclaimer.}
This paper was prepared for informational purposes in part by the Artificial Intelligence Research group of JPMorgan Chase \& Co and its affiliates (``JP Morgan''), and is not a product of the Research Department of JP Morgan.  JP Morgan makes no representation and warranty whatsoever and disclaims all liability, for the completeness, accuracy or reliability of the information contained herein.  This document is not intended as investment research or investment advice, or a recommendation, offer or solicitation for the purchase or sale of any security, financial instrument, financial product or service, or to be used in any way for evaluating the merits of participating in any transaction, and shall not constitute a solicitation under any jurisdiction or to any person, if such solicitation under such jurisdiction or to such person would be unlawful.
\end{acks}

\bibliographystyle{ACM-Reference-Format}
\bibliography{reference-base}


\begin{thebibliography}{45}


\ifx \showCODEN    \undefined \def \showCODEN     #1{\unskip}     \fi
\ifx \showDOI      \undefined \def \showDOI       #1{#1}\fi
\ifx \showISBNx    \undefined \def \showISBNx     #1{\unskip}     \fi
\ifx \showISBNxiii \undefined \def \showISBNxiii  #1{\unskip}     \fi
\ifx \showISSN     \undefined \def \showISSN      #1{\unskip}     \fi
\ifx \showLCCN     \undefined \def \showLCCN      #1{\unskip}     \fi
\ifx \shownote     \undefined \def \shownote      #1{#1}          \fi
\ifx \showarticletitle \undefined \def \showarticletitle #1{#1}   \fi
\ifx \showURL      \undefined \def \showURL       {\relax}        \fi
\providecommand\bibfield[2]{#2}
\providecommand\bibinfo[2]{#2}
\providecommand\natexlab[1]{#1}
\providecommand\showeprint[2][]{arXiv:#2}

\bibitem[\protect\citeauthoryear{Battaglia, Hamrick, Bapst,
  et~al\mbox{.}}{Battaglia et~al\mbox{.}}{2018}]%
        {battaglia2018relational}
\bibfield{author}{\bibinfo{person}{Peter~W Battaglia},
  \bibinfo{person}{Jessica~B Hamrick}, \bibinfo{person}{Victor Bapst},
  {et~al\mbox{.}}} \bibinfo{year}{2018}\natexlab{}.
\newblock \showarticletitle{Relational inductive biases, deep learning, and
  graph networks}.
\newblock \bibinfo{journal}{\emph{arXiv:1806.01261}} (\bibinfo{year}{2018}).
\newblock


\bibitem[\protect\citeauthoryear{Bishop}{Bishop}{2006}]%
        {bishop2006pattern}
\bibfield{author}{\bibinfo{person}{Christopher~M Bishop}.}
  \bibinfo{year}{2006}\natexlab{}.
\newblock \bibinfo{booktitle}{\emph{Pattern recognition and machine learning}}.
\newblock \bibinfo{publisher}{springer}.
\newblock


\bibitem[\protect\citeauthoryear{Brennan and Oliver}{Brennan and
  Oliver}{2013}]%
        {brennan2013emergence}
\bibfield{author}{\bibinfo{person}{Tim Brennan} {and}
  \bibinfo{person}{William~L Oliver}.} \bibinfo{year}{2013}\natexlab{}.
\newblock \showarticletitle{Emergence of machine learning techniques in
  criminology: implications of complexity in our data and in research
  questions}.
\newblock \bibinfo{journal}{\emph{Criminology \& Pub. Pol'y}}
  \bibinfo{volume}{12} (\bibinfo{year}{2013}), \bibinfo{pages}{551}.
\newblock


\bibitem[\protect\citeauthoryear{Chen, Li, Tao, Barnett, Rudin, and Su}{Chen
  et~al\mbox{.}}{2019}]%
        {chen2019looks}
\bibfield{author}{\bibinfo{person}{Chaofan Chen}, \bibinfo{person}{Oscar Li},
  \bibinfo{person}{Daniel Tao}, \bibinfo{person}{Alina Barnett},
  \bibinfo{person}{Cynthia Rudin}, {and} \bibinfo{person}{Jonathan~K Su}.}
  \bibinfo{year}{2019}\natexlab{}.
\newblock \showarticletitle{This looks like that: deep learning for
  interpretable image recognition}. In \bibinfo{booktitle}{\emph{Advances in
  Neural Information Processing Systems (NeurIPS)}}.
  \bibinfo{pages}{8928--8939}.
\newblock


\bibitem[\protect\citeauthoryear{Dhurandhar, Chen, et~al\mbox{.}}{Dhurandhar
  et~al\mbox{.}}{2018}]%
        {dhurandhar2018explanations}
\bibfield{author}{\bibinfo{person}{Amit Dhurandhar}, \bibinfo{person}{Pin-Yu
  Chen}, {et~al\mbox{.}}} \bibinfo{year}{2018}\natexlab{}.
\newblock \showarticletitle{Explanations based on the missing: Towards
  contrastive explanations with pertinent negatives}. In
  \bibinfo{booktitle}{\emph{NeurIPS}}. \bibinfo{pages}{592--603}.
\newblock


\bibitem[\protect\citeauthoryear{Du, Liu, and Hu}{Du et~al\mbox{.}}{2019}]%
        {du2019techniques}
\bibfield{author}{\bibinfo{person}{Mengnan Du}, \bibinfo{person}{Ninghao Liu},
  {and} \bibinfo{person}{Xia Hu}.} \bibinfo{year}{2019}\natexlab{}.
\newblock \showarticletitle{Techniques for interpretable machine learning}.
\newblock \bibinfo{journal}{\emph{Commun. ACM}} \bibinfo{volume}{63},
  \bibinfo{number}{1} (\bibinfo{year}{2019}), \bibinfo{pages}{68--77}.
\newblock


\bibitem[\protect\citeauthoryear{Fonollosa}{Fonollosa}{2019}]%
        {fonollosa2019conditional}
\bibfield{author}{\bibinfo{person}{Jos{\'e}~AR Fonollosa}.}
  \bibinfo{year}{2019}\natexlab{}.
\newblock \showarticletitle{Conditional distribution variability measures for
  causality detection}.
\newblock In \bibinfo{booktitle}{\emph{Cause Effect Pairs in Machine
  Learning}}. \bibinfo{publisher}{Springer}, \bibinfo{pages}{339--347}.
\newblock


\bibitem[\protect\citeauthoryear{Ghorbani, Wexler, Zou, and Kim}{Ghorbani
  et~al\mbox{.}}{2019}]%
        {ghorbani2019towards}
\bibfield{author}{\bibinfo{person}{Amirata Ghorbani}, \bibinfo{person}{James
  Wexler}, \bibinfo{person}{James Zou}, {and} \bibinfo{person}{Been Kim}.}
  \bibinfo{year}{2019}\natexlab{}.
\newblock \showarticletitle{Towards automatic concept-based explanations}.
\newblock \bibinfo{journal}{\emph{arXiv preprint arXiv:1902.03129}}
  (\bibinfo{year}{2019}).
\newblock


\bibitem[\protect\citeauthoryear{Goodfellow, Pouget-Abadie, Mirza, Xu,
  Warde-Farley, et~al\mbox{.}}{Goodfellow et~al\mbox{.}}{2014}]%
        {goodfellow2014generative}
\bibfield{author}{\bibinfo{person}{Ian Goodfellow}, \bibinfo{person}{Jean
  Pouget-Abadie}, \bibinfo{person}{Mehdi Mirza}, \bibinfo{person}{Bing Xu},
  \bibinfo{person}{David Warde-Farley}, {et~al\mbox{.}}}
  \bibinfo{year}{2014}\natexlab{}.
\newblock \showarticletitle{Generative adversarial nets}. In
  \bibinfo{booktitle}{\emph{NeurIPS}}. \bibinfo{pages}{2672--2680}.
\newblock


\bibitem[\protect\citeauthoryear{Goyal, Feder, Shalit, and Kim}{Goyal
  et~al\mbox{.}}{2019a}]%
        {goyal2019explaining}
\bibfield{author}{\bibinfo{person}{Yash Goyal}, \bibinfo{person}{Amir Feder},
  \bibinfo{person}{Uri Shalit}, {and} \bibinfo{person}{Been Kim}.}
  \bibinfo{year}{2019}\natexlab{a}.
\newblock \showarticletitle{Explaining classifiers with causal concept effect
  (cace)}.
\newblock \bibinfo{journal}{\emph{arXiv:1907.07165}} (\bibinfo{year}{2019}).
\newblock


\bibitem[\protect\citeauthoryear{Goyal, Wu, Ernst, et~al\mbox{.}}{Goyal
  et~al\mbox{.}}{2019b}]%
        {goyal2019counterfactual}
\bibfield{author}{\bibinfo{person}{Yash Goyal}, \bibinfo{person}{Ziyan Wu},
  \bibinfo{person}{Jan Ernst}, {et~al\mbox{.}}}
  \bibinfo{year}{2019}\natexlab{b}.
\newblock \showarticletitle{Counterfactual Visual Explanations}. In
  \bibinfo{booktitle}{\emph{International Conference on Machine Learning
  (ICML)}}. \bibinfo{pages}{2376--2384}.
\newblock


\bibitem[\protect\citeauthoryear{Griffiths, Chater, Kemp, Perfors, and
  Tenenbaum}{Griffiths et~al\mbox{.}}{2010}]%
        {griffiths2010probabilistic}
\bibfield{author}{\bibinfo{person}{Thomas~L Griffiths}, \bibinfo{person}{Nick
  Chater}, \bibinfo{person}{Charles Kemp}, \bibinfo{person}{Amy Perfors}, {and}
  \bibinfo{person}{Joshua~B Tenenbaum}.} \bibinfo{year}{2010}\natexlab{}.
\newblock \showarticletitle{Probabilistic models of cognition: Exploring
  representations and inductive biases}.
\newblock \bibinfo{journal}{\emph{Trends in cognitive sciences}}
  \bibinfo{volume}{14}, \bibinfo{number}{8} (\bibinfo{year}{2010}),
  \bibinfo{pages}{357--364}.
\newblock


\bibitem[\protect\citeauthoryear{Hall}{Hall}{2007}]%
        {hall2007structural}
\bibfield{author}{\bibinfo{person}{Ned Hall}.} \bibinfo{year}{2007}\natexlab{}.
\newblock \showarticletitle{Structural equations and causation}.
\newblock \bibinfo{journal}{\emph{Philosophical Studies}}
  \bibinfo{volume}{132}, \bibinfo{number}{1} (\bibinfo{year}{2007}),
  \bibinfo{pages}{109--136}.
\newblock


\bibitem[\protect\citeauthoryear{Hoyer, Janzing, Mooij, Peters,
  et~al\mbox{.}}{Hoyer et~al\mbox{.}}{2008}]%
        {hoyer2008nonlinear}
\bibfield{author}{\bibinfo{person}{Patrik Hoyer}, \bibinfo{person}{Dominik
  Janzing}, \bibinfo{person}{Joris~M Mooij}, \bibinfo{person}{Jonas Peters},
  {et~al\mbox{.}}} \bibinfo{year}{2008}\natexlab{}.
\newblock \showarticletitle{Nonlinear causal discovery with additive noise
  models}.
\newblock \bibinfo{journal}{\emph{NeurIPS}}  \bibinfo{volume}{21}
  (\bibinfo{year}{2008}), \bibinfo{pages}{689--696}.
\newblock


\bibitem[\protect\citeauthoryear{Jang, Gu, and Poole}{Jang
  et~al\mbox{.}}{2016}]%
        {jang2016categorical}
\bibfield{author}{\bibinfo{person}{Eric Jang}, \bibinfo{person}{Shixiang Gu},
  {and} \bibinfo{person}{Ben Poole}.} \bibinfo{year}{2016}\natexlab{}.
\newblock \showarticletitle{Categorical reparameterization with
  gumbel-softmax}.
\newblock \bibinfo{journal}{\emph{arXiv:1611.01144}} (\bibinfo{year}{2016}).
\newblock


\bibitem[\protect\citeauthoryear{Joshi, Koyejo, Kim, et~al\mbox{.}}{Joshi
  et~al\mbox{.}}{2018}]%
        {joshi2018xgems}
\bibfield{author}{\bibinfo{person}{Shalmali Joshi}, \bibinfo{person}{Oluwasanmi
  Koyejo}, \bibinfo{person}{Been Kim}, {et~al\mbox{.}}}
  \bibinfo{year}{2018}\natexlab{}.
\newblock \showarticletitle{xGEMs: Generating examplars to explain black-box
  models}.
\newblock \bibinfo{journal}{\emph{arXiv:1806.08867}} (\bibinfo{year}{2018}).
\newblock


\bibitem[\protect\citeauthoryear{Kalainathan, Goudet, and Dutta}{Kalainathan
  et~al\mbox{.}}{2020}]%
        {kalainathan2020causal}
\bibfield{author}{\bibinfo{person}{Diviyan Kalainathan},
  \bibinfo{person}{Olivier Goudet}, {and} \bibinfo{person}{Ritik Dutta}.}
  \bibinfo{year}{2020}\natexlab{}.
\newblock \showarticletitle{Causal Discovery Toolbox: Uncovering causal
  relationships in Python.}
\newblock \bibinfo{journal}{\emph{JMLR}} \bibinfo{volume}{21},
  \bibinfo{number}{37} (\bibinfo{year}{2020}), \bibinfo{pages}{1--5}.
\newblock


\bibitem[\protect\citeauthoryear{Karimi, Barthe, Sch{\"o}lkopf, and
  Valera}{Karimi et~al\mbox{.}}{2020}]%
        {karimi2020survey}
\bibfield{author}{\bibinfo{person}{Amir-Hossein Karimi},
  \bibinfo{person}{Gilles Barthe}, \bibinfo{person}{Bernhard Sch{\"o}lkopf},
  {and} \bibinfo{person}{Isabel Valera}.} \bibinfo{year}{2020}\natexlab{}.
\newblock \showarticletitle{A survey of algorithmic recourse: definitions,
  formulations, solutions, and prospects}.
\newblock \bibinfo{journal}{\emph{arXiv:2010.04050}} (\bibinfo{year}{2020}).
\newblock


\bibitem[\protect\citeauthoryear{K{\"a}stner}{K{\"a}stner}{2011}]%
        {kastner2011umbrella}
\bibfield{author}{\bibinfo{person}{Johannes K{\"a}stner}.}
  \bibinfo{year}{2011}\natexlab{}.
\newblock \showarticletitle{Umbrella sampling}.
\newblock \bibinfo{journal}{\emph{Wiley Interdisciplinary Reviews:
  Computational Molecular Science}} \bibinfo{volume}{1}, \bibinfo{number}{6}
  (\bibinfo{year}{2011}), \bibinfo{pages}{932--942}.
\newblock


\bibitem[\protect\citeauthoryear{Kim, Khanna, and Koyejo}{Kim
  et~al\mbox{.}}{2016}]%
        {kim2016examples}
\bibfield{author}{\bibinfo{person}{Been Kim}, \bibinfo{person}{Rajiv Khanna},
  {and} \bibinfo{person}{Oluwasanmi~O Koyejo}.}
  \bibinfo{year}{2016}\natexlab{}.
\newblock \showarticletitle{Examples are not enough, learn to criticize!
  criticism for interpretability}. In \bibinfo{booktitle}{\emph{NeurIPS}}.
  \bibinfo{pages}{2280--2288}.
\newblock


\bibitem[\protect\citeauthoryear{Kim, Wattenberg, Gilmer, Cai,
  et~al\mbox{.}}{Kim et~al\mbox{.}}{2018}]%
        {kim2018interpretability}
\bibfield{author}{\bibinfo{person}{Been Kim}, \bibinfo{person}{Martin
  Wattenberg}, \bibinfo{person}{Justin Gilmer}, \bibinfo{person}{Carrie Cai},
  {et~al\mbox{.}}} \bibinfo{year}{2018}\natexlab{}.
\newblock \showarticletitle{Interpretability Beyond Feature Attribution:
  Quantitative Testing with Concept Activation Vectors (TCAV)}. In
  \bibinfo{booktitle}{\emph{ICML}}. \bibinfo{pages}{2668--2677}.
\newblock


\bibitem[\protect\citeauthoryear{Koh and Liang}{Koh and Liang}{2017}]%
        {koh2017understanding}
\bibfield{author}{\bibinfo{person}{Pang~Wei Koh} {and} \bibinfo{person}{Percy
  Liang}.} \bibinfo{year}{2017}\natexlab{}.
\newblock \showarticletitle{Understanding black-box predictions via influence
  functions}. In \bibinfo{booktitle}{\emph{ICML}}. \bibinfo{pages}{1885--1894}.
\newblock


\bibitem[\protect\citeauthoryear{Litjens, Kooi, Bejnordi, Setio, Ciompi,
  et~al\mbox{.}}{Litjens et~al\mbox{.}}{2017}]%
        {litjens2017survey}
\bibfield{author}{\bibinfo{person}{Geert Litjens}, \bibinfo{person}{Thijs
  Kooi}, \bibinfo{person}{Babak~Ehteshami Bejnordi}, \bibinfo{person}{Arnaud
  Arindra~Adiyoso Setio}, \bibinfo{person}{Francesco Ciompi}, {et~al\mbox{.}}}
  \bibinfo{year}{2017}\natexlab{}.
\newblock \showarticletitle{A survey on deep learning in medical image
  analysis}.
\newblock \bibinfo{journal}{\emph{Medical image analysis}}
  \bibinfo{volume}{42} (\bibinfo{year}{2017}), \bibinfo{pages}{60--88}.
\newblock


\bibitem[\protect\citeauthoryear{Lundberg and Lee}{Lundberg and Lee}{2017}]%
        {lundberg2017unified}
\bibfield{author}{\bibinfo{person}{Scott~M Lundberg} {and}
  \bibinfo{person}{Su-In Lee}.} \bibinfo{year}{2017}\natexlab{}.
\newblock \showarticletitle{A unified approach to interpreting model
  predictions}. In \bibinfo{booktitle}{\emph{NeurIPS}}.
  \bibinfo{pages}{4765--4774}.
\newblock


\bibitem[\protect\citeauthoryear{Mahajan, Tan, and Sharma}{Mahajan
  et~al\mbox{.}}{2019}]%
        {mahajan2019preserving}
\bibfield{author}{\bibinfo{person}{Divyat Mahajan}, \bibinfo{person}{Chenhao
  Tan}, {and} \bibinfo{person}{Amit Sharma}.} \bibinfo{year}{2019}\natexlab{}.
\newblock \showarticletitle{Preserving causal constraints in counterfactual
  explanations for machine learning classifiers}.
\newblock \bibinfo{journal}{\emph{arXiv:1912.03277}} (\bibinfo{year}{2019}).
\newblock


\bibitem[\protect\citeauthoryear{McClelland}{McClelland}{1992}]%
        {mcclelland1992interaction}
\bibfield{author}{\bibinfo{person}{JL McClelland}.}
  \bibinfo{year}{1992}\natexlab{}.
\newblock \bibinfo{title}{The interaction of nature and nurture in development:
  A parallel distributed processing perspective (Parallel Distributed
  Processing and Cognitive Neuroscience PDP. CNS. 92.6)}.
\newblock
\newblock


\bibitem[\protect\citeauthoryear{Mirza and Osindero}{Mirza and
  Osindero}{2014}]%
        {mirza2014conditional}
\bibfield{author}{\bibinfo{person}{Mehdi Mirza} {and} \bibinfo{person}{Simon
  Osindero}.} \bibinfo{year}{2014}\natexlab{}.
\newblock \showarticletitle{Conditional generative adversarial nets}.
\newblock \bibinfo{journal}{\emph{arXiv:1411.1784}} (\bibinfo{year}{2014}).
\newblock


\bibitem[\protect\citeauthoryear{Mitchell}{Mitchell}{1980}]%
        {mitchell1980need}
\bibfield{author}{\bibinfo{person}{Tom~M Mitchell}.}
  \bibinfo{year}{1980}\natexlab{}.
\newblock \bibinfo{booktitle}{\emph{The need for biases in learning
  generalizations}}.
\newblock \bibinfo{publisher}{Department of Computer Science, Laboratory for
  Computer Science Research}.
\newblock


\bibitem[\protect\citeauthoryear{Moore, Hammerla, and Watkins}{Moore
  et~al\mbox{.}}{2019}]%
        {moore2019explaining}
\bibfield{author}{\bibinfo{person}{Jonathan Moore}, \bibinfo{person}{Nils
  Hammerla}, {and} \bibinfo{person}{Chris Watkins}.}
  \bibinfo{year}{2019}\natexlab{}.
\newblock \showarticletitle{Explaining deep learning models with constrained
  adversarial examples}. In \bibinfo{booktitle}{\emph{Pacific Rim International
  Conference on Artificial Intelligence}}. Springer, \bibinfo{pages}{43--56}.
\newblock


\bibitem[\protect\citeauthoryear{Mothilal, Sharma, and Tan}{Mothilal
  et~al\mbox{.}}{2020}]%
        {mothilal2020explaining}
\bibfield{author}{\bibinfo{person}{Ramaravind~K Mothilal},
  \bibinfo{person}{Amit Sharma}, {and} \bibinfo{person}{Chenhao Tan}.}
  \bibinfo{year}{2020}\natexlab{}.
\newblock \showarticletitle{Explaining machine learning classifiers through
  diverse counterfactual explanations}. In
  \bibinfo{booktitle}{\emph{Proceedings on Fairness, Accountability, and
  Transparency (FAccT)}}. \bibinfo{pages}{607--617}.
\newblock


\bibitem[\protect\citeauthoryear{Park, Mohammadi, Gorde, Jajodia, Park, and
  Kim}{Park et~al\mbox{.}}{2018}]%
        {park2018data}
\bibfield{author}{\bibinfo{person}{Noseong Park}, \bibinfo{person}{Mahmoud
  Mohammadi}, \bibinfo{person}{Kshitij Gorde}, \bibinfo{person}{Sushil
  Jajodia}, \bibinfo{person}{Hongkyu Park}, {and} \bibinfo{person}{Youngmin
  Kim}.} \bibinfo{year}{2018}\natexlab{}.
\newblock \showarticletitle{Data synthesis based on generative adversarial
  networks}.
\newblock \bibinfo{journal}{\emph{arXiv:1806.03384}} (\bibinfo{year}{2018}).
\newblock


\bibitem[\protect\citeauthoryear{Pawelczyk, Broelemann, and Kasneci}{Pawelczyk
  et~al\mbox{.}}{2020}]%
        {pawelczyk2020learning}
\bibfield{author}{\bibinfo{person}{Martin Pawelczyk}, \bibinfo{person}{Klaus
  Broelemann}, {and} \bibinfo{person}{Gjergji Kasneci}.}
  \bibinfo{year}{2020}\natexlab{}.
\newblock \showarticletitle{Learning Model-Agnostic Counterfactual Explanations
  for Tabular Data}. In \bibinfo{booktitle}{\emph{Proceedings of The Web
  Conference 2020}}. \bibinfo{pages}{3126--3132}.
\newblock


\bibitem[\protect\citeauthoryear{Ribeiro, Singh, and Guestrin}{Ribeiro
  et~al\mbox{.}}{2016}]%
        {ribeiro2016should}
\bibfield{author}{\bibinfo{person}{Marco~Tulio Ribeiro},
  \bibinfo{person}{Sameer Singh}, {and} \bibinfo{person}{Carlos Guestrin}.}
  \bibinfo{year}{2016}\natexlab{}.
\newblock \showarticletitle{" Why should i trust you?" Explaining the
  predictions of any classifier}. In \bibinfo{booktitle}{\emph{Proceedings of
  the 22nd ACM SIGKDD conference on knowledge discovery and data mining}}.
  \bibinfo{pages}{1135--1144}.
\newblock


\bibitem[\protect\citeauthoryear{Ribeiro, Singh, and Guestrin}{Ribeiro
  et~al\mbox{.}}{2018}]%
        {ribeiro2018anchors}
\bibfield{author}{\bibinfo{person}{Marco~Tulio Ribeiro},
  \bibinfo{person}{Sameer Singh}, {and} \bibinfo{person}{Carlos Guestrin}.}
  \bibinfo{year}{2018}\natexlab{}.
\newblock \showarticletitle{Anchors: High-precision model-agnostic
  explanations}. In \bibinfo{booktitle}{\emph{32nd AAAI on Artificial
  Intelligence}}.
\newblock


\bibitem[\protect\citeauthoryear{Rissland}{Rissland}{1991}]%
        {rissland1991example}
\bibfield{author}{\bibinfo{person}{Edwina~L Rissland}.}
  \bibinfo{year}{1991}\natexlab{}.
\newblock \showarticletitle{Example-based reasoning}.
\newblock \bibinfo{journal}{\emph{Informal reasoning in education}}
  (\bibinfo{year}{1991}), \bibinfo{pages}{187--208}.
\newblock


\bibitem[\protect\citeauthoryear{Selvaraju, Cogswell, et~al\mbox{.}}{Selvaraju
  et~al\mbox{.}}{2017}]%
        {selvaraju2017grad}
\bibfield{author}{\bibinfo{person}{Ramprasaath~R Selvaraju},
  \bibinfo{person}{Michael Cogswell}, {et~al\mbox{.}}}
  \bibinfo{year}{2017}\natexlab{}.
\newblock \showarticletitle{Grad-Cam: Visual explanations from deep networks
  via gradient-based localization}. In \bibinfo{booktitle}{\emph{Proceedings of
  the IEEE international conference on computer vision (ICCV)}}.
  \bibinfo{pages}{618--626}.
\newblock


\bibitem[\protect\citeauthoryear{Sundararajan, Taly, and Yan}{Sundararajan
  et~al\mbox{.}}{2017}]%
        {sundararajan2017axiomatic}
\bibfield{author}{\bibinfo{person}{Mukund Sundararajan}, \bibinfo{person}{Ankur
  Taly}, {and} \bibinfo{person}{Qiqi Yan}.} \bibinfo{year}{2017}\natexlab{}.
\newblock \showarticletitle{Axiomatic attribution for deep networks}. In
  \bibinfo{booktitle}{\emph{Proceedings of the 34th International Conference on
  Machine Learning-Volume 70}}. JMLR. org, \bibinfo{pages}{3319--3328}.
\newblock


\bibitem[\protect\citeauthoryear{Vermeire and Martens}{Vermeire and
  Martens}{2020}]%
        {vermeire2020explainable}
\bibfield{author}{\bibinfo{person}{Tom Vermeire} {and} \bibinfo{person}{David
  Martens}.} \bibinfo{year}{2020}\natexlab{}.
\newblock \showarticletitle{Explainable Image Classification with Evidence
  Counterfactual}.
\newblock \bibinfo{journal}{\emph{arXiv:2004.07511}} (\bibinfo{year}{2020}).
\newblock


\bibitem[\protect\citeauthoryear{Wachter et~al\mbox{.}}{Wachter
  et~al\mbox{.}}{2017}]%
        {wachter2017counterfactual}
\bibfield{author}{\bibinfo{person}{Sandra Wachter} {et~al\mbox{.}}}
  \bibinfo{year}{2017}\natexlab{}.
\newblock \showarticletitle{Counterfactual explanations without opening the
  black box: Automated decisions and the GDPR}.
\newblock \bibinfo{journal}{\emph{Harv. JL \& Tech.}}  \bibinfo{volume}{31}
  (\bibinfo{year}{2017}), \bibinfo{pages}{841}.
\newblock


\bibitem[\protect\citeauthoryear{White and Garcez}{White and Garcez}{2019}]%
        {white2019measurable}
\bibfield{author}{\bibinfo{person}{Adam White} {and}
  \bibinfo{person}{Artur~d'Avila Garcez}.} \bibinfo{year}{2019}\natexlab{}.
\newblock \showarticletitle{Measurable counterfactual local explanations for
  any classifier}.
\newblock \bibinfo{journal}{\emph{arXiv preprint arXiv:1908.03020}}
  (\bibinfo{year}{2019}).
\newblock


\bibitem[\protect\citeauthoryear{Xu, Skoularidou, Cuesta-Infante, and
  Veeramachaneni}{Xu et~al\mbox{.}}{2019}]%
        {xu2019modeling}
\bibfield{author}{\bibinfo{person}{Lei Xu}, \bibinfo{person}{Maria
  Skoularidou}, \bibinfo{person}{Alfredo Cuesta-Infante}, {and}
  \bibinfo{person}{Kalyan Veeramachaneni}.} \bibinfo{year}{2019}\natexlab{}.
\newblock \showarticletitle{Modeling tabular data using conditional gan}. In
  \bibinfo{booktitle}{\emph{NeurIPS}}. \bibinfo{pages}{7335--7345}.
\newblock


\bibitem[\protect\citeauthoryear{Yang, Liu, Du, et~al\mbox{.}}{Yang
  et~al\mbox{.}}{2021}]%
        {yang2021generative}
\bibfield{author}{\bibinfo{person}{Fan Yang}, \bibinfo{person}{Ninghao Liu},
  \bibinfo{person}{Mengnan Du}, {et~al\mbox{.}}}
  \bibinfo{year}{2021}\natexlab{}.
\newblock \showarticletitle{Generative Counterfactuals for Neural Networks via
  Attribute-Informed Perturbation}.
\newblock \bibinfo{journal}{\emph{arXiv:2101.06930}} (\bibinfo{year}{2021}).
\newblock


\bibitem[\protect\citeauthoryear{Yang, Liu, Du, Zhou, Ji, and Hu}{Yang
  et~al\mbox{.}}{2020}]%
        {yang2020deep}
\bibfield{author}{\bibinfo{person}{Fan Yang}, \bibinfo{person}{Ninghao Liu},
  \bibinfo{person}{Mengnan Du}, \bibinfo{person}{Kaixiong Zhou},
  \bibinfo{person}{Shuiwang Ji}, {and} \bibinfo{person}{Xia Hu}.}
  \bibinfo{year}{2020}\natexlab{}.
\newblock \showarticletitle{Deep Neural Networks with Knowledge Instillation}.
  In \bibinfo{booktitle}{\emph{SDM}}. \bibinfo{pages}{370--378}.
\newblock


\bibitem[\protect\citeauthoryear{Yeh, Kim, et~al\mbox{.}}{Yeh
  et~al\mbox{.}}{2020}]%
        {yeh2020completeness}
\bibfield{author}{\bibinfo{person}{Chih-Kuan Yeh}, \bibinfo{person}{Been Kim},
  {et~al\mbox{.}}} \bibinfo{year}{2020}\natexlab{}.
\newblock \showarticletitle{On Completeness-aware Concept-Based Explanations in
  Deep Neural Networks}.
\newblock \bibinfo{journal}{\emph{NeurIPS}} (\bibinfo{year}{2020}).
\newblock


\bibitem[\protect\citeauthoryear{Zhou, Sun, Bau, and Torralba}{Zhou
  et~al\mbox{.}}{2018}]%
        {zhou2018interpretable}
\bibfield{author}{\bibinfo{person}{Bolei Zhou}, \bibinfo{person}{Yiyou Sun},
  \bibinfo{person}{David Bau}, {and} \bibinfo{person}{Antonio Torralba}.}
  \bibinfo{year}{2018}\natexlab{}.
\newblock \showarticletitle{Interpretable basis decomposition for visual
  explanation}. In \bibinfo{booktitle}{\emph{ECCV}}. \bibinfo{pages}{119--134}.
\newblock


\end{thebibliography}

\appendix

\section{Proof of Theorem~\ref{theo1}}\label{proof1}

Through the umbrella sampling process, we aim to reconstruct the target distribution $\mathcal{P}_{\mathbf{q}}$ with $N$ biased distributions $\mathcal{P}_{\mathbf{q}}^{i}$ by a weighted sum manner, where $i=1,\cdots, N$. In the following, we demonstrate how to derive the corresponding weight vector $\mathbf{w}$ with the aid of the pre-defined overlap matrix $\mathbf{M}$ in Eq.~\ref{ovlp_mx}. 

Without loss of generality, we here assume $\mathbf{q}$ only contains continuous attributes. By adding the umbrella profile $u_{i}$, the biased distribution $\mathcal{P}_{\mathbf{q}}^{i}$ can then be indicated by:
\begin{equation}
    \mathcal{P}_{\mathbf{q}}^{i} = \left. u_{i} \cdot \mathcal{P}_{\mathbf{q}} \right/ w_{i},
\end{equation} 
where $w_{i}$ denotes the normalized weight for $\mathcal{P}_{\mathbf{q}}^{i}$ we aim to obtain. Since $\int \mathcal{P}_{\mathbf{q}}^{i} \ \mathrm{d}\mathbf{q}=1$ always holds for each biased distribution, we can further represent $w_{i}$ as follows:
\begin{equation}\label{weighti}
    w_{i}=\int u_{i} \mathcal{P}_{\mathbf{q}} \ \mathrm{d}\mathbf{q} = \left<u_{i}\right>_{\mathcal{P}_{\mathbf{q}}} .
\end{equation}
To evaluate the generator $G$ over $\mathcal{P}_{\mathbf{q}}$, we have
\begin{equation} \label{th1_main}
\begin{split}
\left< G \right>_{\mathcal{P}_{\mathbf{q}}} & = \int G \mathcal{P}_{\mathbf{q}} \ \mathrm{d}\mathbf{q} = \int G \ \frac{\sum_{i=1}^{N} u_{i}/w_{i}}{\sum_{j=1}^{N} u_{j}/w_{j}} \ \mathcal{P}_{\mathbf{q}} \ \mathrm{d}\mathbf{q}  \\
& = \sum_{i=1}^{N} \int \!\! \frac{G}{\sum_{j=1}^{N} u_{j}/w_{j}} \frac{u_{i}}{w_{i}} \ \mathcal{P}_{\mathbf{q}} \ \mathrm{d}\mathbf{q}  \\
& = \sum_{i=1}^{N} \int \!\! \frac{G}{\sum_{j=1}^{N} u_{j}/w_{j}} \ \mathcal{P}_{\mathbf{q}}^{i} \ \mathrm{d}\mathbf{q} = \sum_{i=1}^{N} \left<\frac{G}{\sum_{j=1}^{N} u_{j}/w_{j}}\right>_{\mathcal{P}_{\mathbf{q}}^{i}}.
\end{split}
\end{equation}
Thus, we know that the original evaluation of $G$ over $\mathcal{P}_{\mathbf{q}}$ can be possibly conducted over the sum of $N$ biased distributions with proper weights. Here, the distribution sum operation can be simply achieved by the direct sampling over each $\mathcal{P}_{\mathbf{q}}^{i}$, where $i=1,\cdots, N$. 

To finally obtain the weights, we utilize the overlap matrix $\mathbf{M}$ defined in Eq.~\ref{ovlp_mx}. Within $\mathbf{M}$, it is noted that $M_{ij}$ equals to $0$ if there is no overlap between $\mathcal{P}_{\mathbf{q}}^{i}$ and $u_{j}$. Further, we can derive the product of the weight vector $\mathbf{w}$ and the $j$-th column of $\mathbf{M}$ as below, based on Eq.~\ref{weighti} and Eq.~\ref{th1_main}.
\begin{equation}
    \sum_{i=1}^{N} w_{i} \cdot M_{ij}=\sum_{i=1}^{N} \left<\frac{u_{j}}{\sum_{k=1}^{N}u_{k}/w_{k}}\right>_{\mathcal{P}_{\mathbf{q}}^{i}}
    =\left<u_{j}\right>_{\mathcal{P}_{\mathbf{q}}}=w_{j}.
\end{equation}
Now, considering all the columns of $\mathbf{M}$, we can then obtain the weight vector $\mathbf{w}$ by solving the equation $\mathbf{w} \mathbf{M} = \mathbf{w}$. This finalize the proof of the Theorem~\ref{theo1}.

\section{Proof of Theorem~\ref{theo2}}\label{proof2}

For simplicity without loss of generality, we here only consider the case with causal dependence between two attributes, which is illustrated by Fig.~\ref{fig:se_mib}. According to the causal graph of $A \rightarrow B$, we can express the corresponding structural equations as below:
\begin{equation}\label{se_proof}
    \left\{
    \begin{array}{ll}
         B = F_{B}(A, Z_{B}) \\
         A \indep Z_{B}, \ A \sim \mathcal{P}_{A}, \ Z_{B} \sim \mathcal{P}_{Z_{B}}
    \end{array}
    \right.
    ,
\end{equation} 
where $Z_{B}$ generally indicates the disturbance term of the structural equation for $B$. However, the generated samples from Eq.~\ref{se_proof} may not necessarily be identified with the causality $A \rightarrow B$ purely from the observational perspective, since there could exist a symmetric case for $B \rightarrow A$ which derives the same joint distribution of $A$ and $B$. Take the following two set of structural equations for example:
\begin{equation}\label{se_example}
    \left\{
    \begin{array}{ll}
         B = \alpha A + Z_{B}, \ A \indep Z_{B}  \\
         A \sim \mathcal{N}(\mu_{A},\sigma_{A}^{2})  \\ 
         Z_{B} \sim \mathcal{N}(\mu_{Z_{B}},\sigma_{Z_{B}}^{2})
    \end{array}
    \right.
    \left\{
    \begin{array}{ll}
         A = \beta B + Z_{A}, \ B \indep Z_{A}  \\
         B \sim \mathcal{N}(\mu_{B},\sigma_{B}^{2})  \\ 
         Z_{A} \sim \mathcal{N}(\mu_{Z_{A}},\sigma_{Z_{A}}^{2})
    \end{array}
    \right.
    ,
\end{equation}
where $\alpha,\beta$ are constants, and $\mu,\sigma$ respectively denotes the mean and standard deviation of the normal distribution $\mathcal{N}$. Now, assuming we have the following relationships satisfied: 
\begin{equation}
    \left\{
    \begin{array}{ll}
         \beta = \left.\alpha \sigma_{A}^{2} \right/ \alpha^{2} \sigma_{A}^{2} + \sigma_{Z_{B}}^{2} \\
         \mu_{B} = \alpha \mu_{A} + \mu_{Z_{B}}, \ \sigma_{B}^{2} = \alpha^{2} \sigma_{A}^{2} + \sigma_{Z_{B}}^{2} \\ 
         \mu_{Z_{A}} = (1-\alpha\beta) \mu_{A} - \beta \mu_{Z_{B}}, \ \sigma_{Z_{A}}^{2} = (1-\alpha\beta)^{2} \sigma_{A}^{2} + \beta^{2} \sigma_{Z_{B}}^{2}
    \end{array}
    \right.\nonumber
    ,
\end{equation}
the two set of structural equations in Eq.~\ref{se_example} then derives the exact same joint distribution for $A$ and $B$, from which we cannot identify the related causal dependence with the generated samples. 

Thus, to guarantee the causality we incorporate is able to be identified, we need to break the symmetry of potential structural equations. One promising way for our case is to utilize the properties of additive noise model (ANM)~\cite{hoyer2008nonlinear}, which is proved to be able to generate the samples with related causal dependence identifiable. To prove a set of structural equations follow ANM, there are two key requirements for verification: (1) the transformation function $F$ is non-linear; (2) the influence of the disturbance term can be isolated out of $F$ in an additive manner. In the following, we verify our design in Theorem~\ref{theo2} is consistent with the ANM formulation. 

\begin{itemize}
  \item[$\blacksquare$] \textit{\textbf{Non-linear Transformation}}: In our design, the transformation $F$ is implemented with feed-forward modules, which consists of several dense layers for computation. As a result, the implemented transformation is thus a non-linear mapping. Specifically for Eq.~\ref{se_proof}, $F_{B}$ is non-linear in our scenario. 
  
  \item[$\blacksquare$] \textit{\textbf{Additive Disturbance}}: For our case, the structural equation in Eq.~\ref{se_proof} can be expressed as: $B=F_{B}(A \oplus Z_{B})$, where $\oplus$ represents the direct sum, indicating the concatenation of $A$ and $Z_{B}$ before feeding into $F_{B}$. Since $A \indep Z_{B}$, we know that there exists relevant transformations $F_{B}^{\prime}$ and $F_{B}^{\prime\prime}$, such that
  \begin{equation}\label{ds_homo}
      B=F_{B}(A \oplus Z_{B})=F_{B}^{\prime}(A)+F_{B}^{\prime\prime}(Z_{B}),
  \end{equation}
  based on the homomorphism property of direct sum. In Eq.~\ref{ds_homo}, $F_{B}^{\prime\prime}(Z_{B})$ can be further treated as a transformed disturbance. Thus, within our scenario, we see that Eq.~\ref{se_proof} can isolate the influence of disturbance $Z_{B}$ as an additive term. 
\end{itemize}

\noindent Overall, we show that our design of $G$ in Theorem~\ref{theo2} essentially builds a series of ANMs for counterfactual generation. With the identifiable causal dependence guaranteed by ANM, we finalize the proof of Theorem~\ref{theo2}.

\section{Details \& Analysis of Umbrella Sampling Settings}\label{appendix_us} 

We utilize the umbrella sampling to obtain the training queries with rare values, aiming to effectively train the proposed MCS and let it well capture the potential hypothetical distributions. During implementation, we divide the particular sampling space into $N$ different windows, where $i$-th window is appended with the umbrella profile $u_{i}$. In experiments, our biased samplers are implemented with the Ensemble Sampler\footnote{https://emcee.readthedocs.io/en/stable/user/sampler/\#emcee.EnsembleSampler}, and $N$ is set as $8, 32$ respectively for the \textit{Adult} and \textit{Home\_Credit} dataset. Besides, we employ $8$ walkers for each umbrella-biased sampler, and run $1000$ steps for each walker. The samplers will stop when the maximum Gelman-Rubin estimate falls below the threshold $\zeta=0.01$. 

\begin{figure}[t]
\centering
\includegraphics[width=0.95\columnwidth]{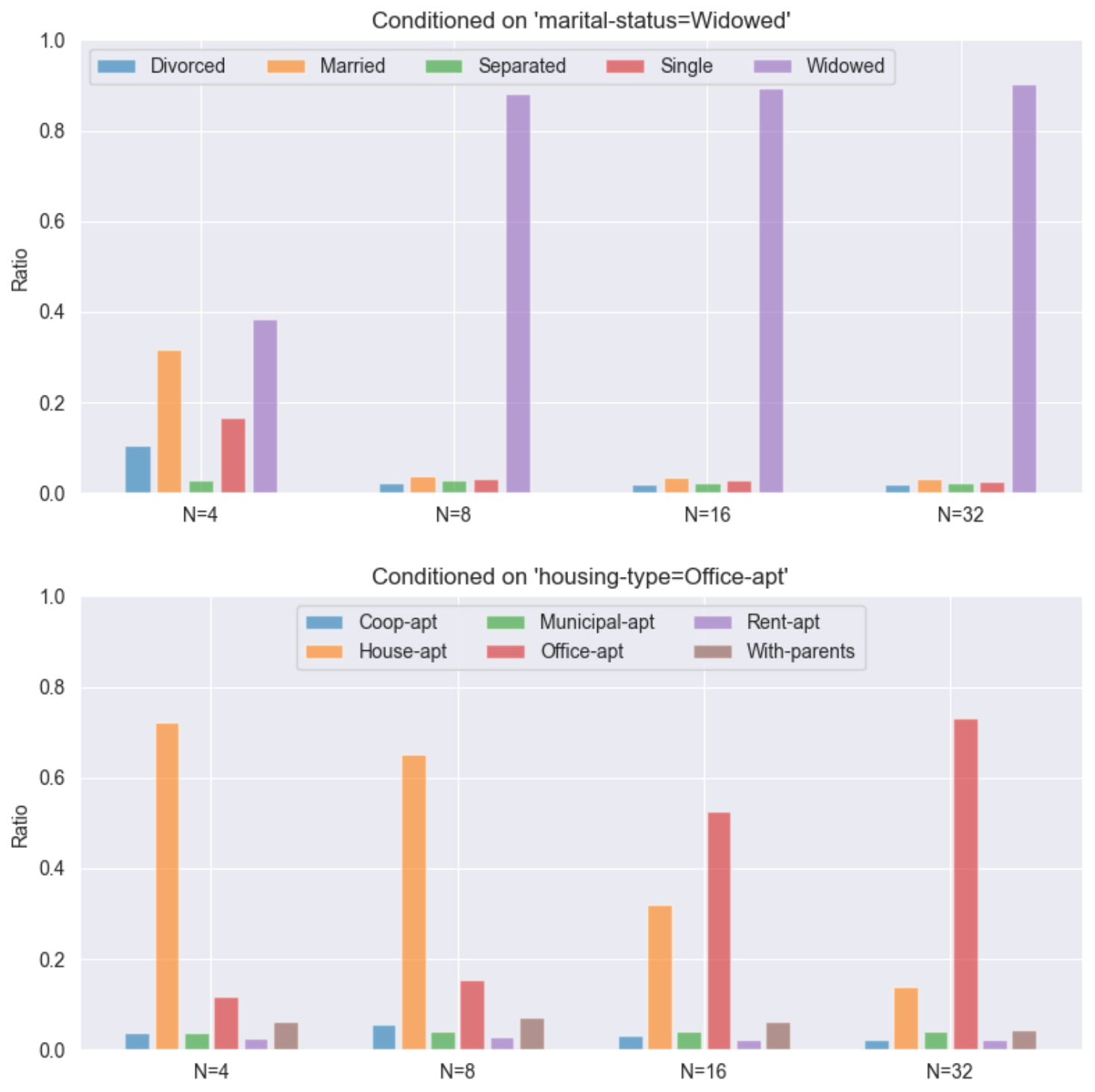}
\vspace{-0.2cm}
\caption{The conditional modeling performance of MCS with umbrella sampling on different number of windows.} 
\label{fig:app_perform}
\end{figure}

\begin{figure}
\centering
\includegraphics[width=\columnwidth]{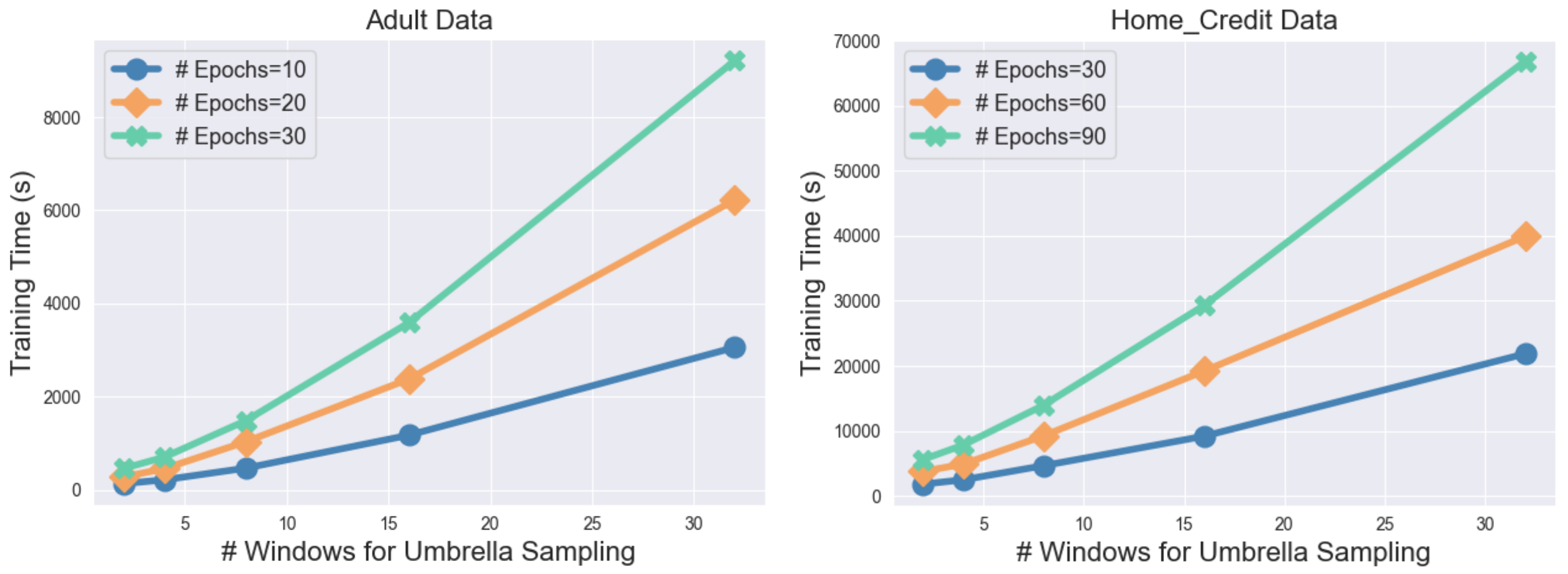}
\vspace{-0.6cm}
\caption{The time consumption of MCS training.} 
\label{fig:app_time}
\end{figure}

Typically, when $N$ is large enough, those rare values would have roughly the equal opportunities to be sampled compared with major ones, which basically balances the values appeared for MCS training. However, it is usually not acceptable with an overlarge $N$, because $N$ also affects the training efficiency significantly. Increasing $N$ will directly lead to the rise of computational complexity, referring Theo.~\ref{theo1}, and will further result in the fact that each training batch consumes more time for updating. Thus, to achieve a reasonable trade-off on training between the effectiveness and efficiency, we need to select an appropriate number of windows $N$ for umbrella sampling. Here, we attach some additional experimental results, considering the influences of $N$, shown in Fig.~\ref{fig:app_perform} and Fig.~\ref{fig:app_time}. According to the comparison, it is noted that MCS trained with more umbrella samplers shows a better modeling performance on hypothetical distributions. Meanwhile, the increase of $N$ brings about heavier time cost for training, which becomes more significant with larger data scale and more training epochs. Therefore, properly choosing $N$ for specific data should be an important pre-step before deploying MCS for interpretation in practice. 

\section{Synthesizer Training with Different Deployed Classifiers}\label{appendix_loss} 

\begin{figure}
\centering
\includegraphics[width=\columnwidth]{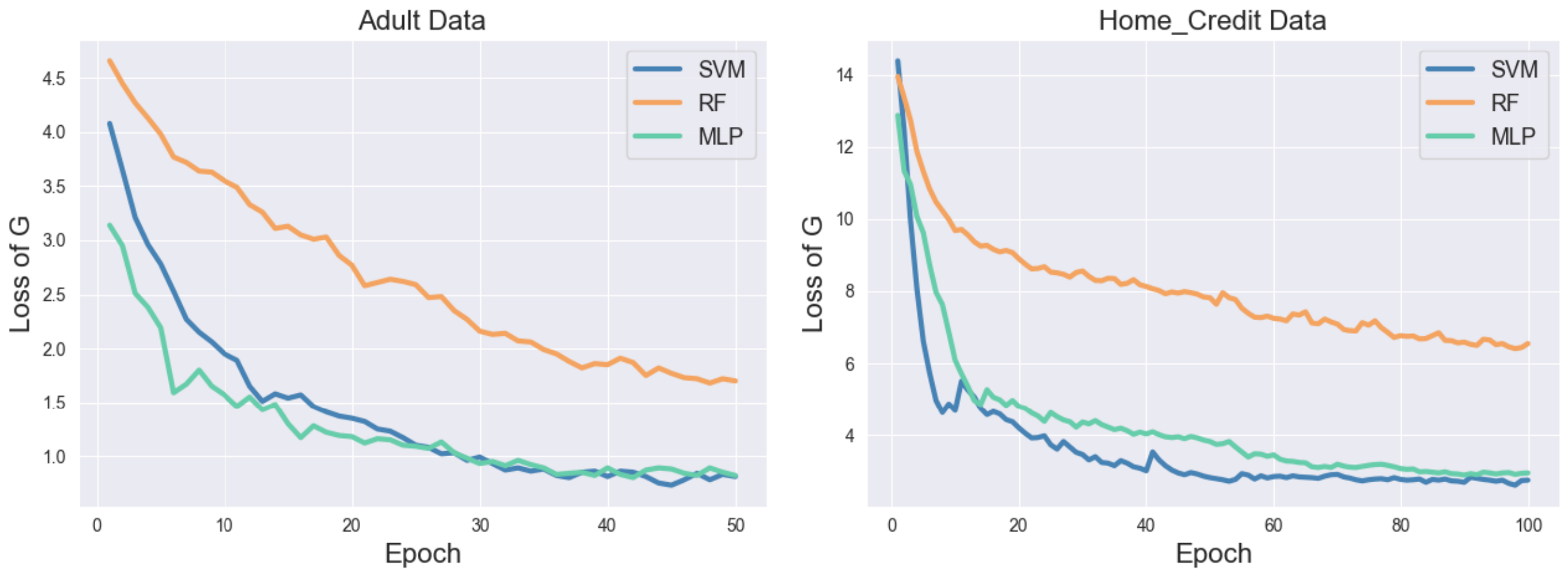}
\vspace{-0.6cm}
\caption{The training loss of generator $G$ with different $f$.} 
\label{fig:app_loss}
\end{figure}

We show some additional results about the MCS training regarding to different deployed classifiers $f$ (i.e., SVM, RF, MLP), and briefly discuss how $f$ can potentially affect the training process. In experiments, the dimension of the latent space (i.e., $\mathbf{z}$) is set as $128$, and $D, G$ are designed with two-layer feed-forward modules whose intermediate dimensions are set as $256$. Besides, the training batch size is fixed as $500$, and the learning rate is $2 \times 10^{-4}$. The empirical results on the loss of generator $G$ during training, with respect to different $f$, are illustrated in Fig.~\ref{fig:app_loss}. Based on the training loss curve, we note that generator $G$ converges faster when trained with deployed SVM and MLP, which typically takes less epochs to reach certain loss value compared with that under the RF case. This observation mainly results from the $L^{ce}$ term in Eq.~\ref{eq:cf_loss}, which involves particular $f$ to calculate the counterfactual loss. Thus, when decision boundaries of the deployed $f$ are not smooth, the corresponding $L^{ce}$ term may not be effectively minimized for training objectives, so that it would largely increase the difficulties of the optimizer in updating $G$. In our experiments, the deployed SVM and MLP classifier seem to have smoother boundaries than RF (intuitively validated by Fig.~\ref{fig:cf_gen_bound}), and related training objectives are better optimized within certain epochs. From this perspective, it sheds light on the relationship between the counterfactual synthesizer and the deployed classifier, where smoother $f$ would boost the MCS training towards better effectiveness. 

\section{Dataset Preprocessing}\label{appendix_data} 

In the conducted experiments of this paper, we preprocess the two real-world datasets employed (i.e., \textit{Adult} and \textit{Home\_Credit}) for simplicity. The processing actions include the following aspects:
\begin{itemize}[leftmargin=*]
\item \textit{Remove the rows with missing values.} This action is mainly for the \textit{Home\_Credit} data, which contains many instances with \textit{None}; 

\item \textit{Delete some columns (attributes).} We ignore some columns which may not be that significant for prediction. For example, the `fnlwgt' attribute in \textit{Adult} data is removed in our experiments. 

\item \textit{Merge some values.} We merge those values together which share overlap semantics. For example, in \textit{Adult} data, value \texttt{Assoc-voc}, \texttt{Assoc-acdm} are merged as \texttt{Assoc}, and value \texttt{11th}, $\cdots$, \texttt{1st-4th} are all merged as \texttt{School}, within the attribute `education'. 
\end{itemize}

\end{document}